\documentclass[10pt,twocolumn,letterpaper]{article}

\usepackage{cvpr}              

\usepackage{graphicx}
\usepackage{amsmath}
\usepackage{amssymb}
\usepackage{booktabs}
\usepackage{adjustbox}
\usepackage[accsupp]{axessibility}  
\usepackage{algorithm}
\usepackage{algorithmic}
\usepackage{wrapfig}
\usepackage{xcolor}
\usepackage{enumitem}
\usepackage{mathtools}
\usepackage{adjustbox}
\usepackage{enumitem}
\setitemize{itemsep=0pt,topsep=0pt,parsep=0pt,partopsep=0pt}
\input{Definitions}

\usepackage[pagebackref,breaklinks,colorlinks]{hyperref}

\usepackage[capitalize]{cleveref}
\crefname{section}{Sec.}{Secs.}
\Crefname{section}{Section}{Sections}
\Crefname{table}{Table}{Tables}
\crefname{table}{Tab.}{Tabs.}

\def\cvprPaperID{11850} 
\def\confName{CVPR}
\def\confYear{2023}

\begin{document}

\title{Shifted Diffusion for Text-to-image Generation}

\author{Yufan Zhou $^1$\thanks{Performed this work during internship at ByteDance, code is available at \url{https://github.com/drboog/Shifted_Diffusion}. The research of the first and last author was supported in part by NSF through grants IIS-1910492 and CCF-2200173 and by KAUST CRG10-4663.2.}, 
~
Bingchen Liu $^2$,
~
Yizhe Zhu $^2$,
~
Xiao Yang $^2$,
~
Changyou Chen $^1$,
~
Jinhui Xu $^1$\\

$^1$ State University of New York at Buffalo~~~~ 
$^2$ ByteDance ~~~~\\
{\tt\small \{yufanzho, changyou, jinhui\}@buffalo.edu, \{bingchenliu, yizhe.zhu, yangxiao.0\}@bytedance.com}

}
\maketitle

\begin{abstract}
We present Corgi, a novel method for text-to-image generation. Corgi is based on our proposed shifted diffusion model, which achieves better image embedding generation from input text. Unlike the baseline diffusion model used in DALL-E 2, our method seamlessly encodes prior knowledge of the pre-trained CLIP model in its diffusion process by designing a new initialization distribution and a new transition step of the diffusion. Compared to the strong DALL-E 2 baseline, our method performs better in generating image embedding from the text in terms of both efficiency and effectiveness, resulting in better text-to-image generation. Extensive large-scale experiments are conducted and evaluated in terms of both quantitative measures and human evaluation, indicating a stronger generation ability of our method compared to existing ones. Furthermore, our model enables semi-supervised and language-free training for text-to-image generation, where only part or none of the images in the training dataset have an associated caption. Trained with only $1.7\%$ of the images being captioned, our semi-supervised model obtains FID results comparable to DALL-E 2 on zero-shot text-to-image generation evaluated on MS-COCO. Corgi also achieves new state-of-the-art results across different datasets on downstream language-free text-to-image generation tasks, outperforming the previous method, Lafite, by a large margin.
\end{abstract}

\section{Introduction}
``AI-generated content" has attracted increasingly more public awareness thanks to the significant progress in recent research of high-fidelity text-aligned image synthetic tasks.~\cite{ramesh2021zero,rombach2021high,ramesh2022hierarchical,saharia2022photorealistic,gu2022vector, tang2022improved,yu2022scaling}. Particularly, models trained on web-scale datasets have demonstrated their impressive ability to generate out-of-distribution images from arbitrary text inputs that describe unseen combinations of visual concepts. 

Starting from DALL-E~\cite{ramesh2021zero}, researchers have proposed a variety of approaches to further 
advance the state-of-the-art (SOTA) of text-to-image generation in terms of both generation quality and efficiency.
Latent Diffusion Model~\cite{rombach2021high} trains a diffusion model in the latent space of auto-encoder instead of pixel space, leading to better generation efficiency. GLIDE~\cite{nichol2021glide} adopts a hierarchical architecture, which consists of diffusion models at different resolutions. Such a model design strategy has shown to be effective and has been adopted by many follow-up works.
DALL-E 2~\cite{ramesh2022hierarchical} further introduces an extra image embedding input.
Such an image embedding not only improves the model performance in text-to-image generation but also enables applications, including image-to-image generation and generation under multi-modal conditions.
Imagen~\cite{saharia2022photorealistic} makes use of a rich pre-trained text encoder~\cite{raffel2020exploringt5}, demonstrating that a frozen text encoder pre-trained on the large-scale text-only dataset can help text-to-image generation models in understanding the semantics of text descriptions.
Parti~\cite{yu2022scaling} shows a further successful scale-up of the generative model, leading to impressive improvement in text-to-image consistency with transformer structure. 

The aforementioned approaches focus on improving text-to-image generation by either scaling up trainable/frozen modules or designing better model architectures. In this work, we explore an orthogonal direction, where we propose novel techniques to improve the diffusion process itself and make it more suitable and effective for text-to-image generation.

\begin{figure*}[ht!]
    \centering
    \begin{subfigure}[t]{0.3\linewidth}
        \includegraphics[width=1.\linewidth]{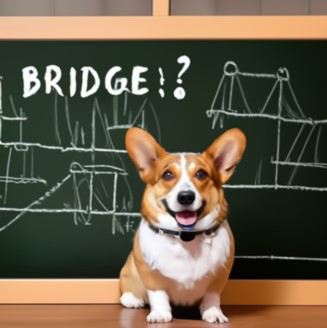}
        \caption{A corgi in front of a blackboard with ``Bridge!" on it.}
    \end{subfigure}
    \begin{subfigure}[t]{0.3\linewidth}
        \includegraphics[width=1.\linewidth]{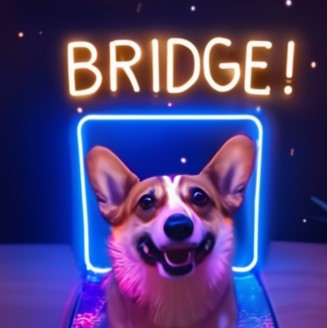}
        \caption{A corgi with a neon light reads ``Bridge!".}
    \end{subfigure}
    \begin{subfigure}[t]{0.3\linewidth}
        \includegraphics[width=1.\linewidth]{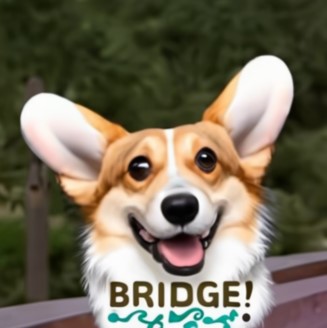}
        \caption{A realistic photo about corgi with word ``Bridge!".}
    \end{subfigure}
    \begin{subfigure}[t]{0.3\linewidth}
        \includegraphics[width=1.\linewidth]{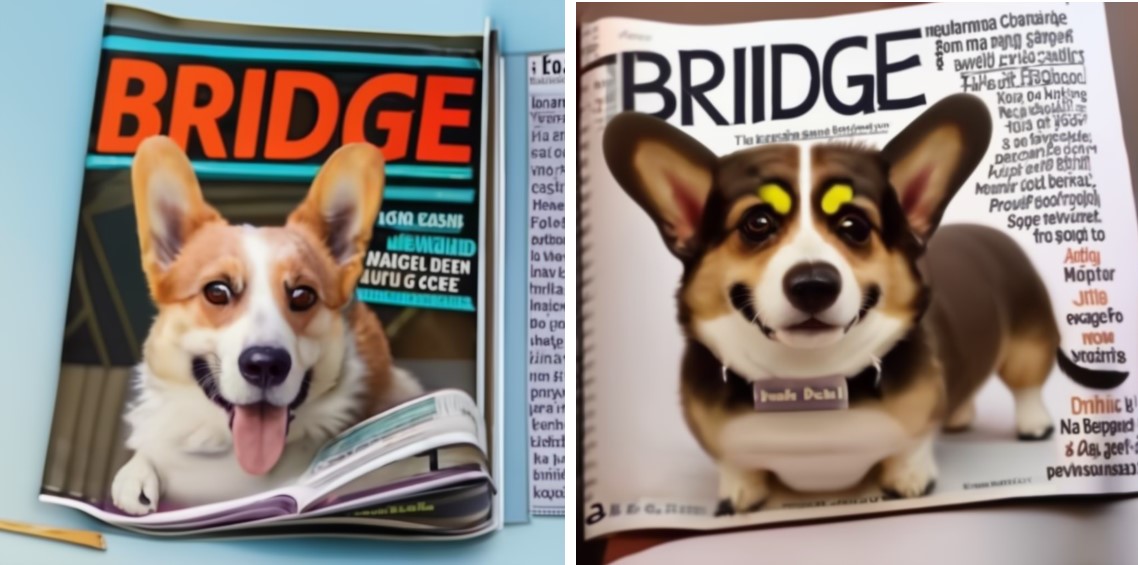}
        \caption{A magazine about Corgi, which has ``Bridge" on it.}
    \end{subfigure}   
    \begin{subfigure}[t]{0.3\linewidth}
        \includegraphics[width=1.\linewidth]{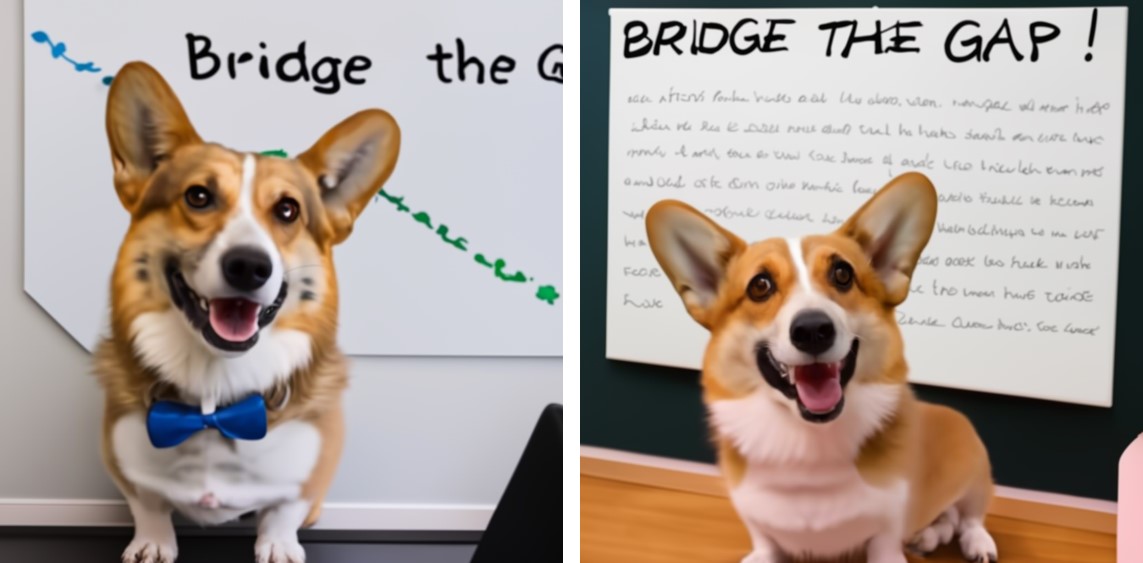}
        \caption{A corgi is smiling, in front of a whiteboard with ``Bridge the Gap" on it.}
    \end{subfigure}
    \begin{subfigure}[t]{0.3\linewidth}
        \includegraphics[width=1.\linewidth]{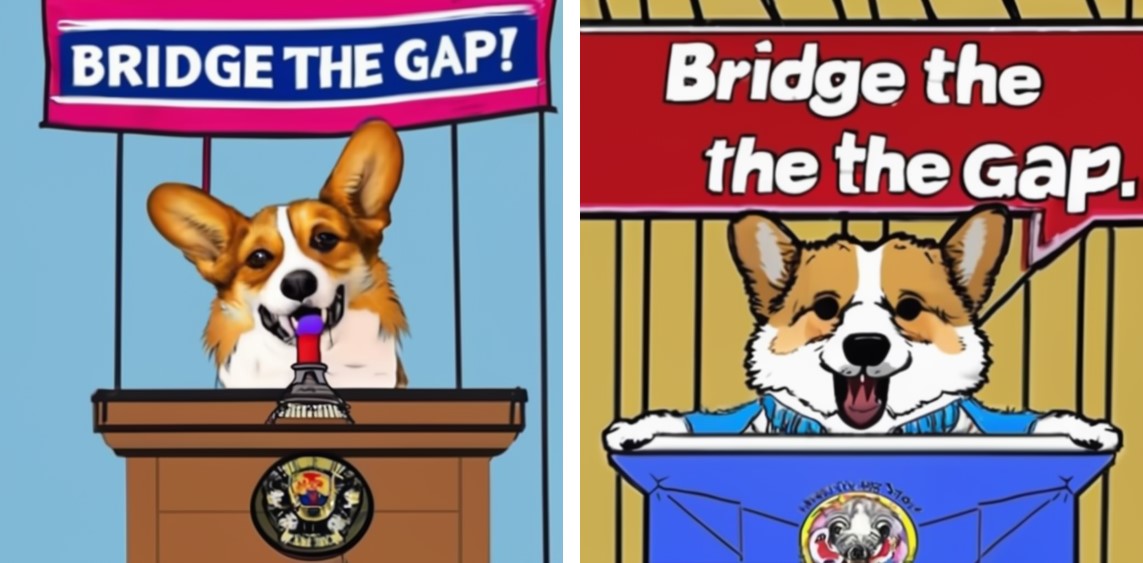}
        \caption{A corgi is giving a speech about ``Bridge the Gap" in cartoon style.}
    \end{subfigure}
    \vspace{-0.15in}
    \caption{We propose Corgi, a novel diffusion model designed for flexible text-to-image generation which can ``bridge the gap". 
    }
    \label{fig:corgi}
    \vspace{-0.2in}
\end{figure*}

Specifically, we propose Corgi (\textbf{C}lip-based shifted diffusi\textbf{O}n model b\textbf{R}id\textbf{GI}ng the gap), a novel diffusion model designed for a flexible text-to-image generation. Our model can perform text-to-image generation under all the supervised, semi-supervised, and language-free settings.
By ``bridging the gap," we emphasize two key novelties in our method: (1) our model tries to bridge the image-text modality gap~\cite{Liang2022MindTG} so as to train a better generative model. Modality gap is a critical concept discovered in pre-trained models such as CLIP~\cite{radford2021learning}, which captures the phenomenon 
that multi-modality representations do not align in the joint embedding space. With Corgi, we can better utilize CLIP in text-to-image generation, which is shown in the experiment to achieve better generation; (2) our model bridges the gap of data availability for different researchers/communities. In our design, Corgi can naturally enable semi-supervised and language-free text-to-image generation, where only a small portion or even none of the images in the training dataset are captioned. This is important because the cost of constructing a high-quality large image-text-paired dataset could be prohibitive, especially in the case where hundreds of millions of images need to be captioned.
We show that by merely using an image-only dataset and the public CC15M dataset~\cite{Sharma2018ConceptualCA,changpinyo2021conceptualcc12m}, Corgi achieves promising results comparable to SOTA models on open-domain text-to-image generation tasks. 
To summarize, our contributions are:
\begin{itemize}[nosep]
    \item We propose Corgi, a novel diffusion model that seamlessly incorporates prior knowledge from the pre-trained model ({\it e.g.,}CLIP) into its diffusion process;
    \item Our method is general and can 
    be applied in 
    different settings of text-to-image generation, {\it e.g.}, it naturally enables semi-supervised and language-free text-to-image generation;
    \item Extensive and large-scale experiments are conducted. Both quantitative and qualitative results illustrate the effectiveness of the proposed method. We especially demonstrate that, with only $1.7\%$ captioned images in the training dataset, it is possible to achieve results comparable to the SOTA method. In addition, our model achieves SOTA results under language-free setting across different downstream datasets.
\end{itemize}

\begin{figure}[t!]
    \centering
    \begin{subfigure}[t]{0.95\linewidth}
        \includegraphics[width=0.95\linewidth]{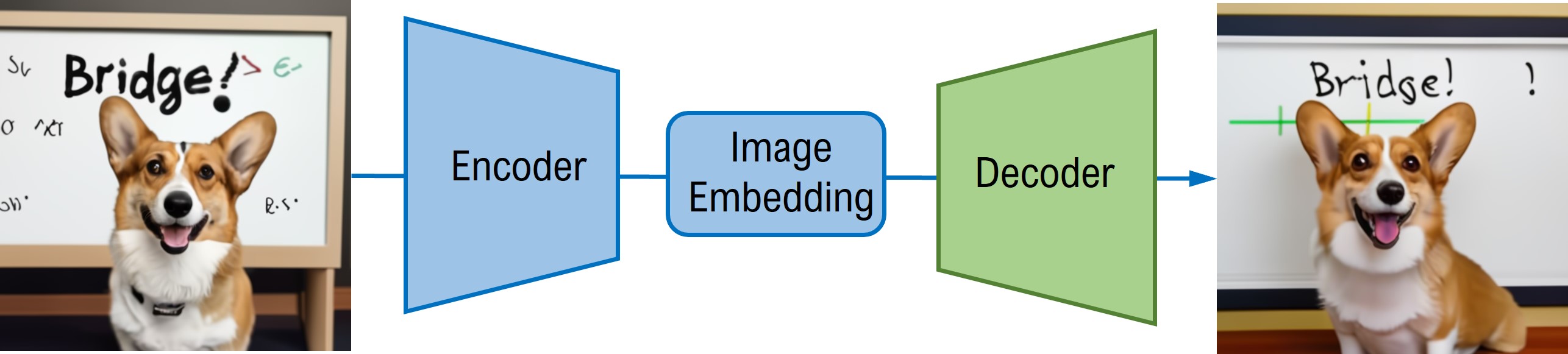}
        \caption{Train a decoder which generates images from image embeddings.}
    \end{subfigure}
    \begin{subfigure}[t]{0.95\linewidth}
        \includegraphics[width=0.95\linewidth]{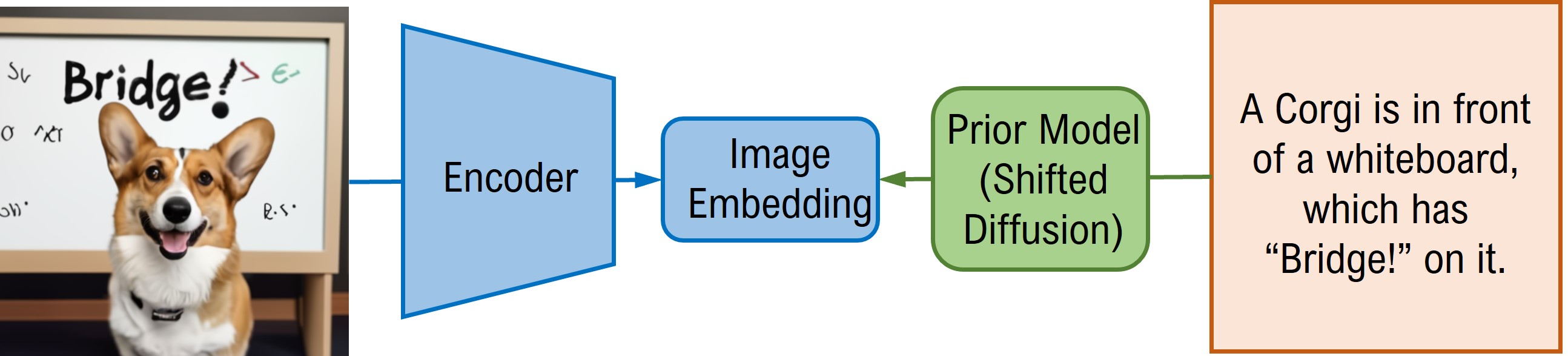}
        \caption{Train a shifted diffusion model which generates embeddings from text.}
    \end{subfigure}   
    \begin{subfigure}[t]{0.95\linewidth}
        \includegraphics[width=0.95\linewidth]{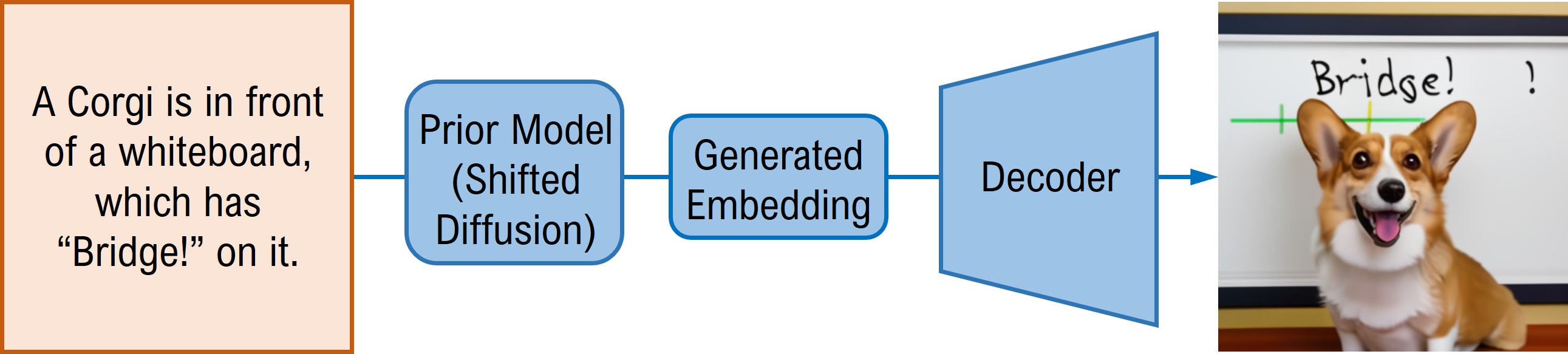}
        \caption{Generate image with text input at inference.}
    \end{subfigure}        
    \vspace{-0.1in}
    \caption{Illustration of our framework. The trainable modules are colored in green, and the frozen modules are colored in blue.}
    \label{fig:model}
    \vspace{-0.3in}
\end{figure}
\section{Methodology}
We start by illustrating the proposed general framework for text-to-image generation. Our framework is shown in Figure \ref{fig:model}, which consists of three key components: (1) a pre-trained image encoder that maps images to their embeddings; (2) a decoder that generates images from the corresponding embeddings; and (3) a prior model that generates image embeddings from the corresponding text captions. In our implementation, we use the pre-trained CLIP image encoder because its output space is a multi-modal embedding space that has been demonstrated to benefit the text-to-image generation task~\cite{zhou2021lafite}. The decoder can be either a diffusion model or a generative adversarial network (GAN). Note that if one chooses the decoder as a hierarchical diffusion model and makes it conditioned on both image embedding and text, our final 
structure will be similar to DALL-E 2~\cite{ramesh2022hierarchical}. 

We adopt this framework because of its flexibility: it can perform different types of generation tasks such as text-to-image generation, image-to-image generation, and generation conditioned on both image and text. In addition, our framework naturally enables semi-supervised training, {\it i.e.}, the training dataset can be a mixture of image-text pairs and pure images that are not captioned. In this setting, the image-text pairs will be used to train the prior model, and all the pure images will be used to train the decoder. Such a semi-supervised training setting is important in practice, especially when training a text-to-image generation model on new domains with a limited budget. 
As discussed in \cite{zhou2021lafite}, constructing high-quality image-text pairs could be very expensive and requires a heavy human workload. Our framework provides the community with the flexibility of choosing the number of images to be captioned based on their budget. As it will be shown in the experiments, semi-supervised training can obtain impressive results, which are even comparable to those of supervised training.

\begin{figure}[t!]
    \centering
    \includegraphics[width=0.5\linewidth]{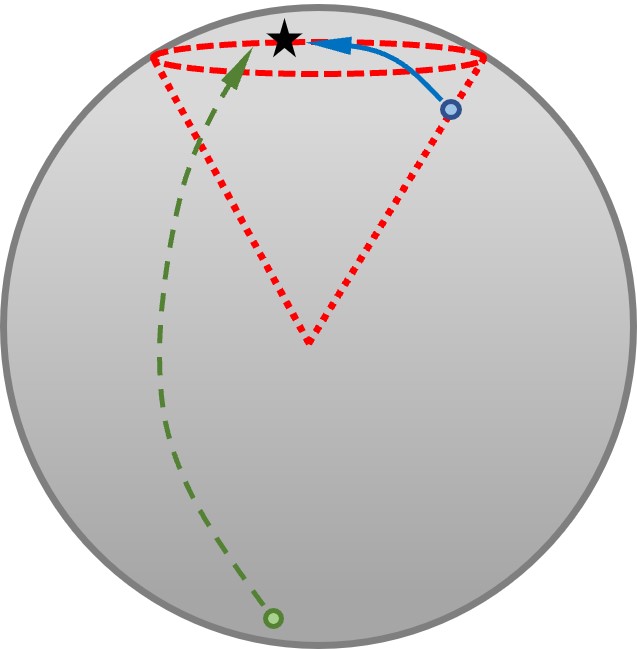}
    \caption{Illustration of diffusions inside the multi-modal joint space of CLIP. We use the cone in red dotted lines to represent the effective output space of CLIP image encoder. The target embedding is represented by the black star. The green dashed arrow represents the sampling process of the baseline diffusion model, whose starting point is random noise. The blue solid arrow represents the sampling process of our shifted diffusion, whose starting point is an image embedding inside the red cone.}
    \label{fig:clip_space}
    \vspace{-0.25in}
\end{figure}

In this paper, we focus on improving the prior model, which is less exploited in previous works.
As shown in DALL-E 2~\cite{ramesh2022hierarchical}, a diffusion-based prior model is introduced to generate the target CLIP image embedding $\zb_0$ via the following sequential sampling process: $\zb_{t-1} \sim p_{\thetab}(\zb_{t-1} \vert \zb_t, \yb)$ for $t=T, ..., 1$, where $\zb_T \sim \mathcal{N}(\mathbf{0}, \mathbf{I})$, $t$ denotes timestep, $\yb$ denotes text caption, and $p_{\thetab}(\cdot|\cdot)$ is the inverse transition distribution induced from the diffusion model. 
Although it is shown in \cite{ramesh2022hierarchical} that this prior model can benefit generation in terms of both image-text alignment and image fidelity, 
we suspect that this vanilla sampling process may not be appropriate for generating high-quality CLIP image embeddings (which will be the inputs of decoder thus greatly influence the generation quality). 
The reason is that, as revealed in \cite{Liang2022MindTG}, the effective output space of the CLIP image encoder is actually a very small region of the whole embedding space, as shown in Figure \ref{fig:clip_space}. 
Consequently,  $\zb_T \sim \mathcal{N}(\mathbf{0}, \mathbf{I})$, which is the starting point of sampling, might be far away from the target embedding.
Intuitively, if $\zb_T$ is closer to the target $\zb_0$, we may be able to well approximate $\zb_0$ within fewer sampling steps. Likewise, we may better approximate the target within the same number of steps, if the initialization is closer to the target.

Based on this motivation, we propose shifted diffusion, a novel diffusion model that considers prior knowledge contained in the pre-trained CLIP image encoder. Specifically, the noise distribution $p(\zb_T)$ of shifted diffusion is a parametric distribution rather than standard Gaussian $\mathcal{N}(\mathbf{0}, \mathbf{I})$.
We present the details of our method in the following.

\subsection{Shifted Diffusion}
Let $q(\zb_0)$ be the distribution of ground-truth image embeddings in the joint latent space of the CLIP model. 
Because of the fact that valid image embeddings only occupy a small region of the whole embedding space~\cite{Liang2022MindTG}, in order to achieve better generation in a diffusion process, 
we would like to construct a new $p(\zb^T)$ which is expected to be more similar to $q(\zb^0)$ than standard Gaussian. 
However, $q(\zb_0)$ is an intractable distribution, making it challenging to train a diffusion model based on it.

To tackle the aforementioned problem, we consider the initial distribution $p(\zb_T)$ to be a parametric Gaussian distribution as $\mathcal{N}(\zb_T; \mub, \Sigmab)$, and can be obtained by simply analyzing the training dataset\footnote{In practice, we set $\Sigmab$ to be a diagonal matrix with its element $\Sigmab_{i,i}= \kappa \sigma_i$. $\sigma_i$ is the standard deviation of $i^{th}$ element of all image embeddings from the training dataset, $\kappa>0$ is a constant for scaling.}. We design the transition $q(\zb_t \vert \zb_{t-1})$ to be a Gaussian as
\begin{align}\label{eq:transition}
    q(\zb_t \vert \zb_{t-1}) \coloneqq \mathcal{N}(\zb_t; \sqrt{1 - \beta_t} \zb_{t-1} + \sbb_t, \beta_t \Sigmab),
\end{align}
where $\beta_t$ is a constant following \cite{ho2020denoising}. 
Compared to the vanilla diffusion with 
\[
q(\zb_t \vert \zb_{t-1}) \coloneqq \mathcal{N}(\zb_t; \sqrt{1-\beta_t} \zb_{t-1}, \beta_t \mathbf{I}),
\]
our diffusion process introduces a shift term $\sbb_t$ at every timestep $t$, thus termed {\em shifted diffusion}.

One can show that $q(\zb_t \vert \zb_0)$ has a closed-form expression (all proofs are provided in the Appendix):
\begin{align*}
    q(\zb_t \vert \zb_0) = \mathcal{N}(\zb_t; \sqrt{\bar{\alpha}_{t}} \zb_0 +  \sum_{i=1}^{t} \sbb_i\sqrt{\bar{\alpha}_{t}/\bar{\alpha}_i}, (1 - \bar{\alpha}_{t})\Sigmab),
\end{align*}
where $\bar{\alpha}_{t} = \prod_{i=1}^t (1-\beta_i)$. 
Specifically, we choose $\sbb_t = (1- \sqrt{1-\beta_t})\mub$, leading to
\begin{align}\label{eq:q_t}
    \hspace{-0.2cm} q(\zb_t \vert \zb_0) = \mathcal{N}(\zb_t; \sqrt{\bar{\alpha}_{t}} \zb_0 + (1 - \sqrt{\bar{\alpha}_{t}})\mub, (1 - \bar{\alpha}_{t})\Sigmab).
\end{align}

\begin{remark}
As we can see, by selecting $\{\beta_t\}_{t=1}^T$ such that $\alpha_T \approx 0$, $q(\zb_T \vert \zb_0) $ can approximate $\mathcal{N}(\zb_T; \mub, \Sigmab)$, the distribution of the image embeddings. In other words, shifted diffusion process is a process that transfers a ground-truth image embedding into a \textit{random image embedding}\footnote{The random image embedding might be scaled as we use a scaling factor $\kappa$.}; whereas the vanilla diffusion process is a process that transfers an image embedding into a random \textit{Gaussian noise}, which is certainly not what we expect. 
\end{remark}
From Equation \eqref{eq:transition} and \eqref{eq:q_t}, we can get the closed-form expression of $q(\zb_t \vert \zb_{t-1}), q(\zb_t \vert \zb_0), q(\zb_{t-1} \vert \zb_0)$.
By the property of Gaussian distribution~\cite{bishop2006PRML} and some simple derivations, we can get the posterior distribution
\begin{align}\label{eq:q_posterior}
    & q(\zb_{t-1} \vert \zb_t, \zb_0) = \mathcal{N}(\zb_{t-1}; \nub, \mathbf{\Lambda}), \\
    & \nub = \gamma (\zb_t - \sbb_t) + \eta \zb_0 + \tau  (1 - \sqrt{\bar{\alpha}_{t-1}})\mub, \nonumber \\
    &\mathbf{\Lambda} = (1 - \bar{\alpha}_{t-1})\beta_t \Sigmab/(1-\bar{\alpha}_{t}), \nonumber 
\end{align}
where 
\vspace{-0.8cm}
\begin{align*}
    &\gamma = (1 - \bar{\alpha}_{t-1})\sqrt{\alpha_t}/(1-\bar{\alpha}_{t}), \\
    & \eta = \beta_t\sqrt{\bar{\alpha}_{t-1}}/(1 - \bar{\alpha}_{t}), \\
    & \tau = \beta_t/(1 - \bar{\alpha}_{t}).
\end{align*}

With Equations \eqref{eq:q_t} and \eqref{eq:q_posterior}, we can easily optimize the loss function of the proposed shifted diffusion 
\vspace{-0.1cm}
\begin{align}\label{eq:diffusion_loss}
    \mathbf{L}_{\thetab} = \mathbb{E}_{q} \{ &\text{D}_{\text{KL}}(q(\zb_T \vert \zb_0) \Vert p(\zb_T)) - \log p_{\thetab} (\zb_0 \vert \zb_1) \\
    & + \sum_{t>1} \text{D}_{\text{KL}}(q(\zb_{t-1} \vert \zb_t, \zb_0) \Vert p_{\thetab}(\zb_{t-1} \vert \zb_t))\}~,\nonumber
\end{align}
\vspace{-0.1cm}
where  $\text{D}_{\text{KL}}$ denotes KL-divergence.
Since both $q(\zb_t \vert \zb_0)$ and $q(\zb_{t-1} \vert \zb_t, \zb_0) $ are Gaussian distributions by our design, the KL-divergence terms have closed-form solutions, enabling easy stochastic optimization. 

\begin{algorithm}[t!]
    \caption{Shifted Diffusion}\label{algo:sft}
    \begin{algorithmic}[1]
        \STATE {\color{blue} // Training}
        \STATE {\bfseries Require: $\{p_i(\zb_T)=\mathcal{N}(\zb_T; \mub_i, \Sigmab_i)\}_{i=1}^k$, a diffusion model parameterized by $\thetab$, CLIP image encoder $f_{\text{img}}$ and text encoder $f_{\text{txt}}$.} 
        \WHILE { {\em not converge} }{   
            \STATE{Sample image-text data pair $(\xb_0, \yb)$, $\zb_0 = f_{\text{img}}(\xb_0)$}
            \STATE{Select corresponding $p_{c_{\yb}}$ by \eqref{eq:assign}}
            \STATE{Update $\thetab$ by gradient descent w.r.t $\text{L}_{\thetab}$}
            \IF{$\mub_i, \Sigmab_i$ are learn-able}
                \STATE{
                Update $\mub_i, \Sigmab_i$ by gradient descent w.r.t $\text{L}_{p}$}
            \ENDIF
        }
        \ENDWHILE
    \STATE{\color{blue} // Inference}
    \STATE{Given a text caption $\yb$, select its corresponding distribution $p_{c_{\yb}}$, sample corresponding $\zb_T$}
    \FOR{t=T,...,1}
    \STATE{Sample $\zb_{t-1} \sim p_{\thetab}(\zb_{t-1} \vert \zb_t)$}
    \ENDFOR
    \STATE{Return $\zb_0$}
    \end{algorithmic}
\end{algorithm}

Although setting $p(\zb_T)$ to be a Gaussian distribution leads to a simplified loss that can be conveniently optimized, it also introduces some drawbacks as the ground-truth distribution of image embedding is not a single-mode Gaussian. 
To tackle this problem, we propose to use a collection of Gaussian distributions, denoted as $\{p_i(\zb_T)\}_{i=1}^k$ with $p_i(\zb_T) \coloneqq \mathcal{N}(\zb_T; \mub_i, \Sigmab_i)$. Let $\zb_0$ and $\yb$ be the ground-truth image embedding and its associated text caption, respectively. For each pair of $(\zb_0, \yb)$, we select its corresponding Gaussian $p_{c_{\yb}}$ by the top-1 cosine similarity as
\vspace{-0.1cm}
\begin{align}\label{eq:assign}
    c_{\yb} = \argmax_{1\leq i \leq k} \text{Sim}(\mub_i, f_{\text{txt}}(\yb)),
\end{align}
\vspace{-0.3cm}
where $\f_{\text{txt}}$ is the pre-trained CLIP text encoder,  and $c_{\yb}$ denotes the index of the selected Gaussian. One may also use
\begin{align}\label{eq:assign_noise}
   c_{\yb} = \argmax_{1\leq i \leq k} \mathbb{E}_{\epsilonb_i \sim p_{i}} \left[\text{Sim}(\epsilonb_i, f_{\text{txt}}(\yb))\right]~, 
\end{align}
which requires more computation because of the expectation. 
After selecting $p_{c_{\yb}}(\zb_T)$, its parameters $\mub_{c_{\yb}}$ and $\Sigmab_{c_{\yb}}$ will be used in Equation \eqref{eq:q_t} and \eqref{eq:q_posterior} for optimization. An extra positional embedding representing $c_{\yb}$ is also injected into the diffusion model, with a similar implementation as the time embedding for $t$. Compared to a single-mode Gaussian distributions, $\{p_i(\zb_T)\}_{i=1}^k$ is supposed to have better expressive ability, and $p_{c_{\yb}}$ is expected to better initialize $\zb_T$ to make it closer to the target $\zb_0$.

In our implementation, $\{p_i(\zb_T)\}_{i=1}^k$ is estimated by performing clustering on the training dataset. Similar to existing quantization methods~\cite{van2017neural}, we can also learn $(\mub_i, \Sigmab_i)$ by optimization. Specifically, we propose to update $\mub_i, \Sigmab_i$ during training with the following loss function
\begin{align}\label{eq:loss_gaussians}
    \mathbf{L}_p =&  - \mathbb{E}_{\zb_0} \{\mathbb{E}_{\zb_T \sim p_{c_{\yb}}(\zb_T)} \{\text{Sim}(\zb_T, \zb_0)\}\} \\
    &+ \xi \dfrac{\sum_{i\neq j}}{k(k-1)}   \mathbb{E}_{\zb_{T} \sim p_i(\zb_T)} \{\mathbb{E}_{\zb^\prime_T \sim p_j(\zb_T)} \{\text{Sim}(\zb_T, \zb_T^\prime)\}\}, \nonumber
\end{align}
where $\xi$ is a hyper-parameter. The first term of $\mathbf{L}_p$ forces the sample $\zb_T$ from Gaussian $p_{c_{\yb}}$ to be close to the corresponding ground-truth $\zb_0$, and the second term ensures that $p_i(\zb_T)$ does not overlap each other too much. Note that $\{p_i(\zb_T)\}_{i=1}^k$ is only optimized with $\mathbf{L}_p$, {\it i.e.}, we manually stop the gradient back-propagation from $\text{L}_{\thetab}$ to $\{\mub_i, \Sigmab_i\}_{i=1}^k$. Our algorithm is summarized in Algorithm \ref{algo:sft}, with more implementation details provided in the experiments section.

\begin{figure*}[ht!]
    \centering
    \begin{subfigure}[t]{0.22\linewidth}
        \includegraphics[width=1.\linewidth]{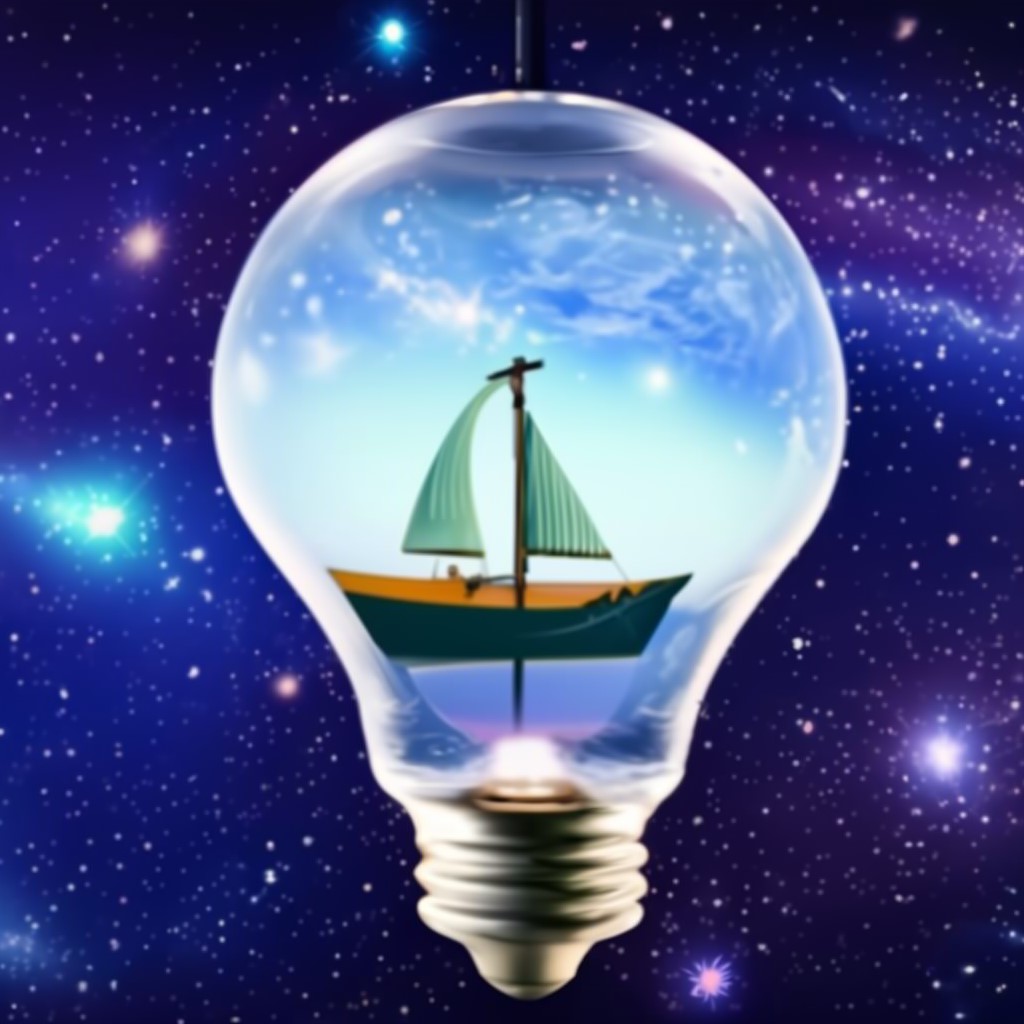}
        \caption{A photo of a light bulb in outer space traveling the galaxy with a sailing boat inside the light bulb.}
    \end{subfigure}       ~
    \begin{subfigure}[t]{0.22\linewidth}
        \includegraphics[width=1.\linewidth]{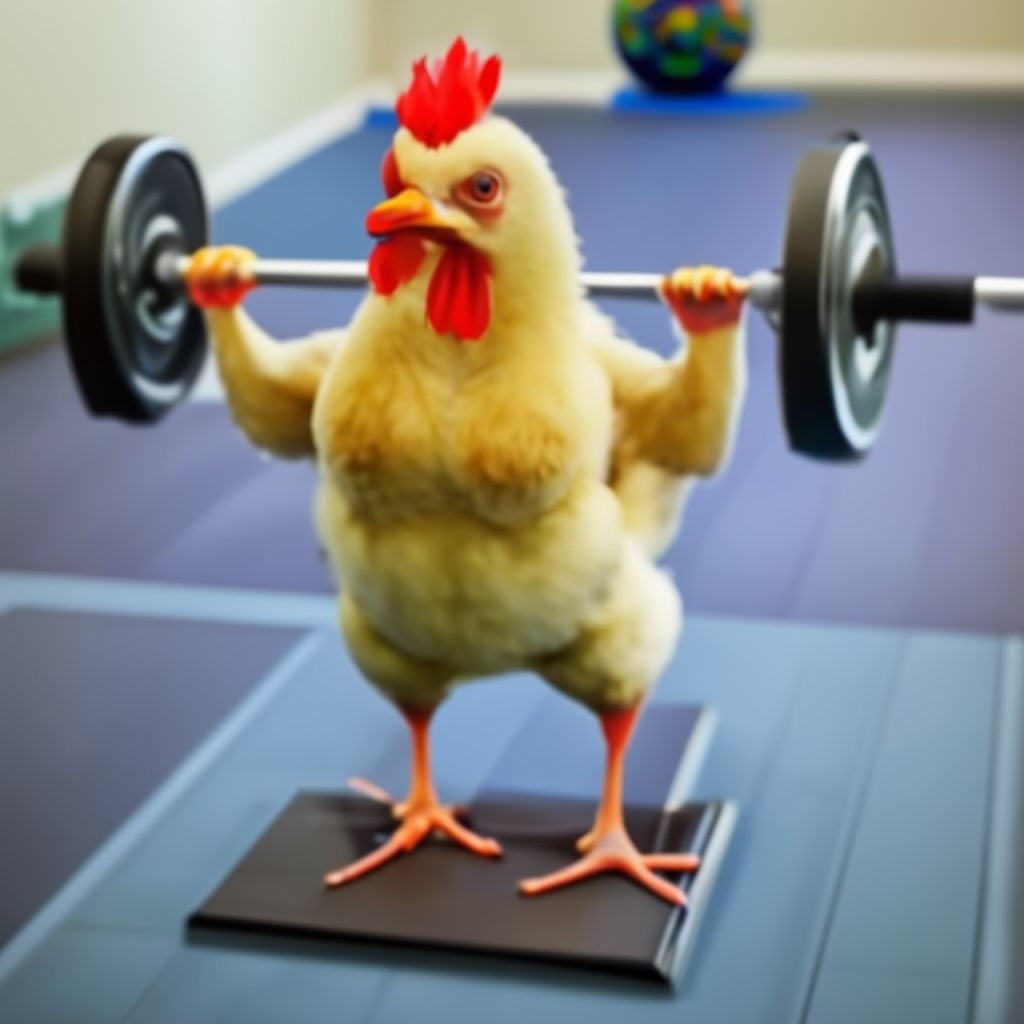}
        \caption{A high-resolution photo of a chicken working out in a gym.}
    \end{subfigure}      ~
    \begin{subfigure}[t]{0.22\linewidth}
        \includegraphics[width=1.\linewidth]{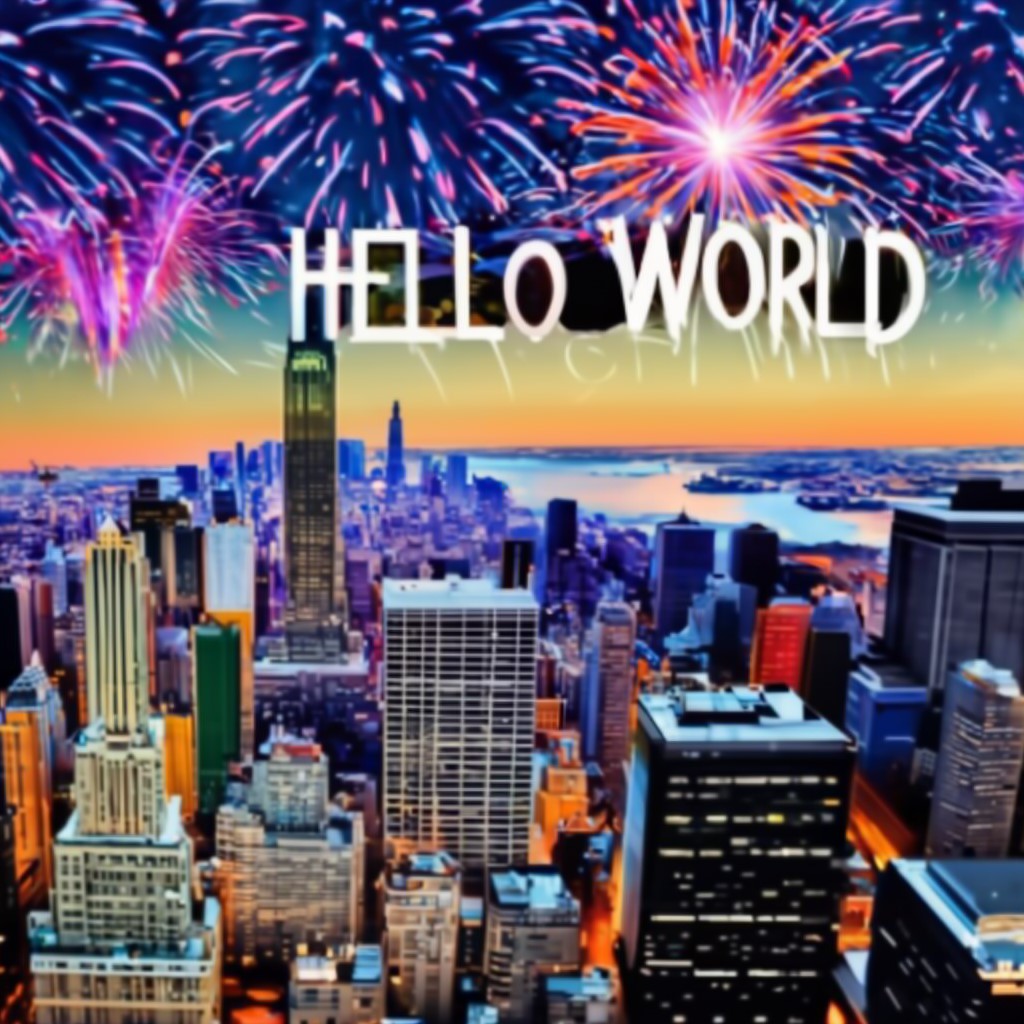}
        \caption{New York Skyline with 'Hello World' written with fireworks on the sky.}
    \end{subfigure}        ~
    \begin{subfigure}[t]{0.22\linewidth}
        \includegraphics[width=1.\linewidth]{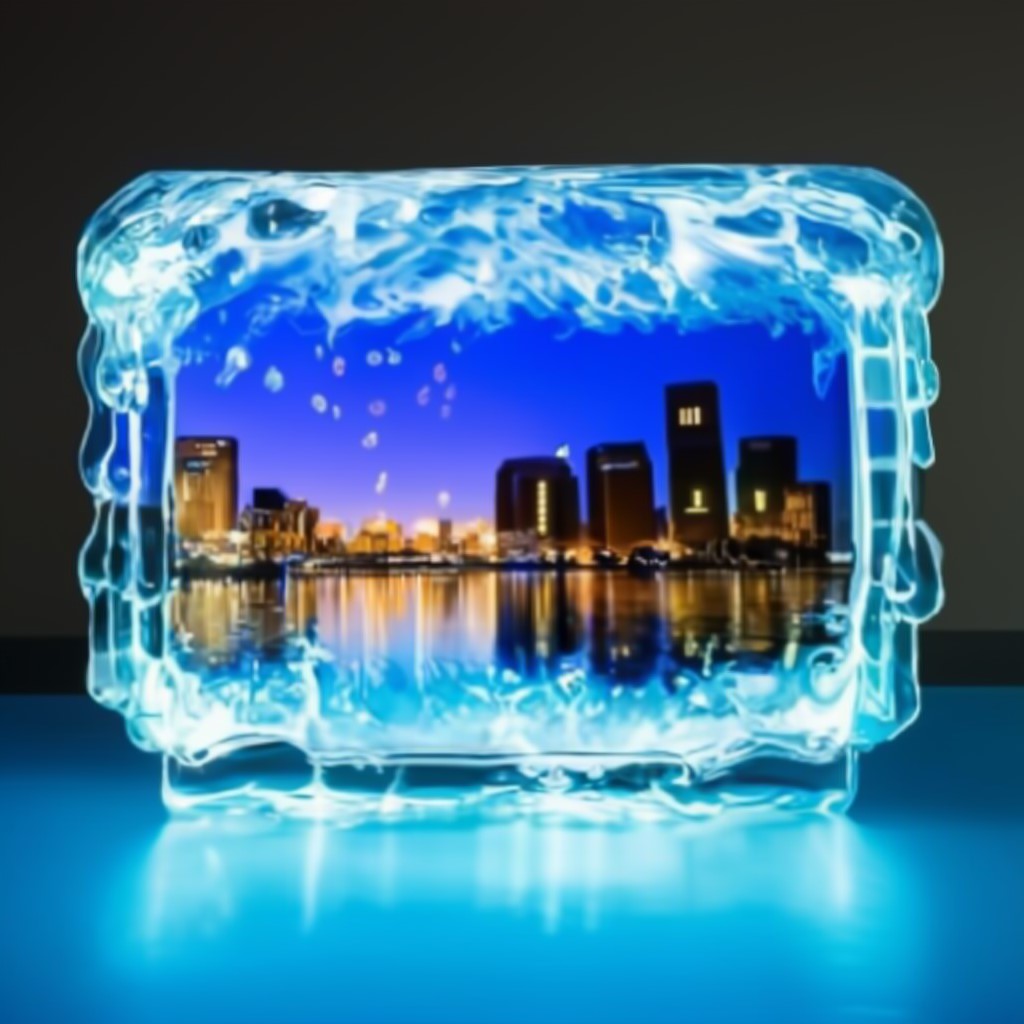}
        \caption{A television made of water that displays an image of a cityscape at night.}
    \end{subfigure}     
    \begin{subfigure}[t]{0.22\linewidth}
        \includegraphics[width=1.\linewidth]{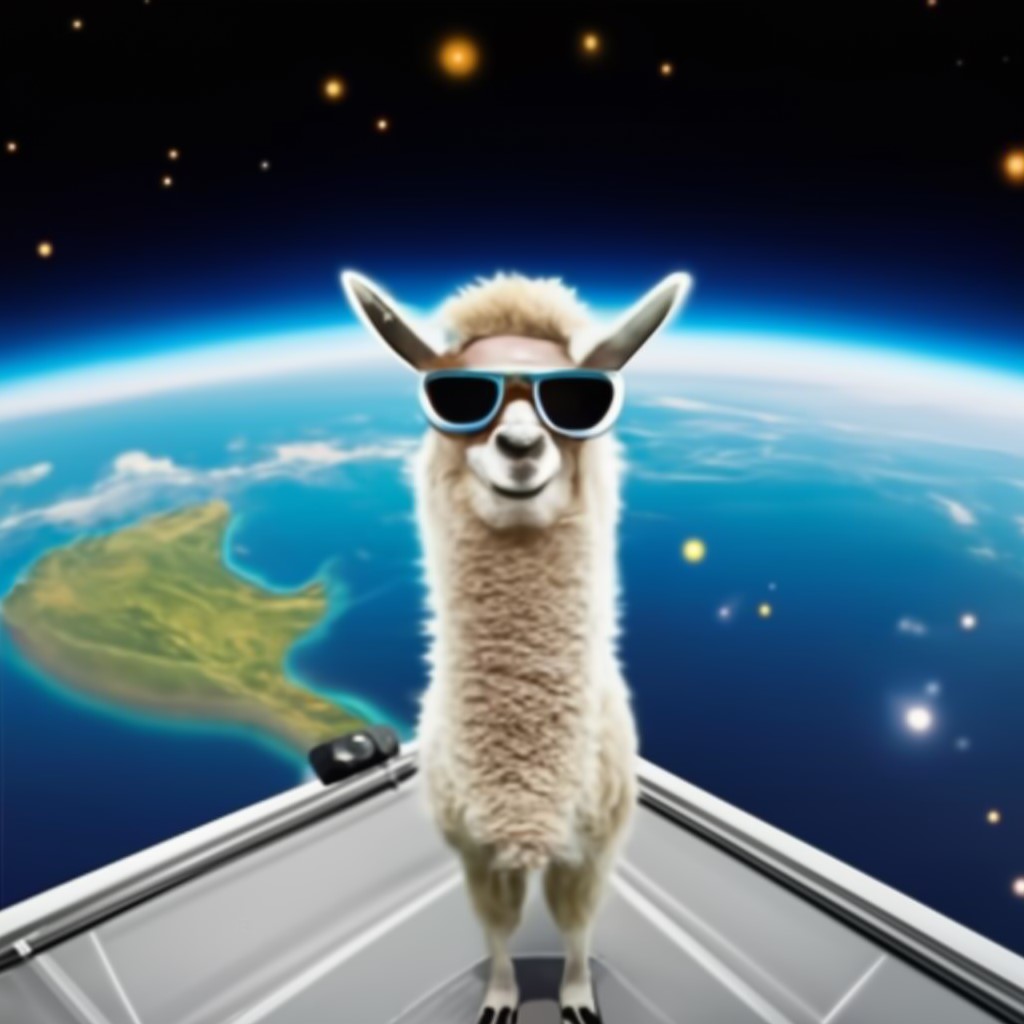}
        \caption{A photo of llama wearing sunglasses standing on the deck of a spaceship with the Earth in the background.}
    \end{subfigure}       ~    
    \begin{subfigure}[t]{0.22\linewidth}
        \includegraphics[width=1.\linewidth]{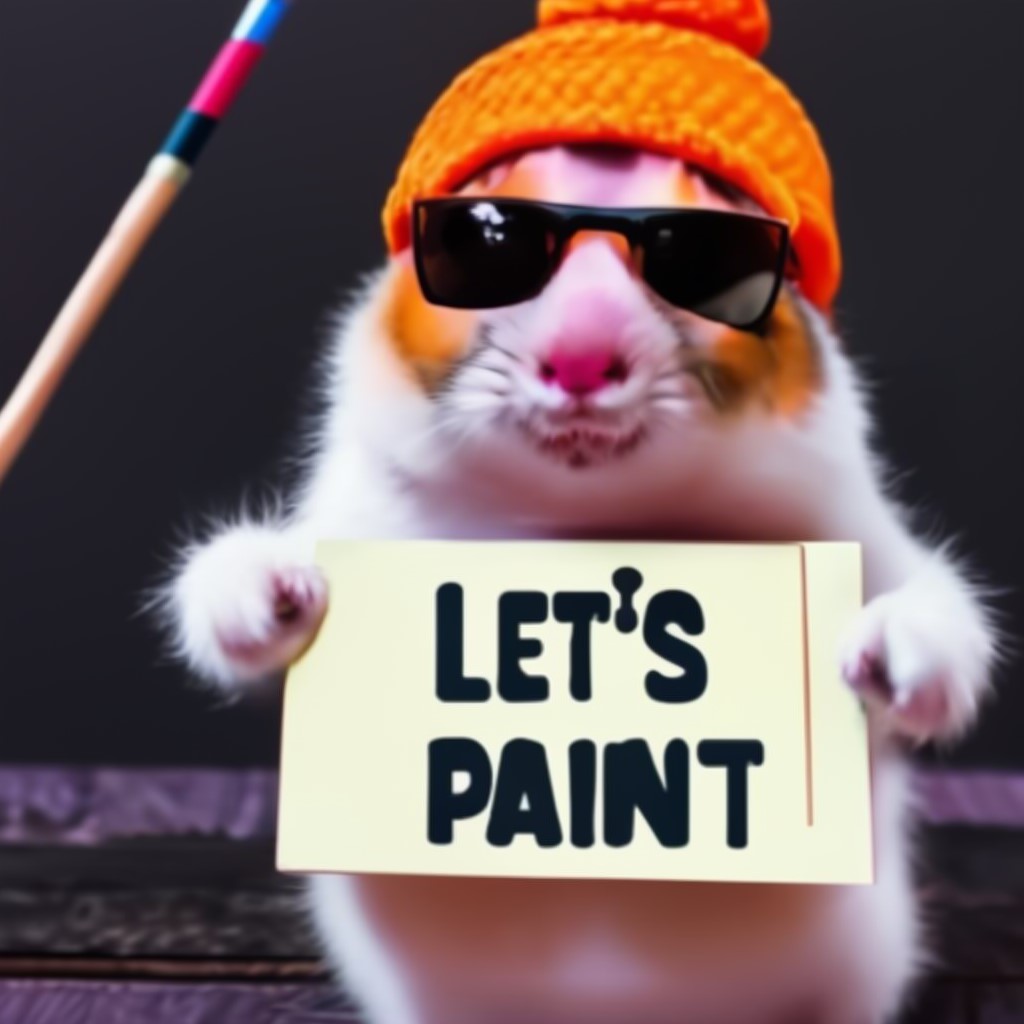}
        \caption{A high contrast portrait photo of a fluffy hamster wearing an orange beanie and sunglasses holding a sign that says ``Let's PAINT!".}
    \end{subfigure}      ~
    \begin{subfigure}[t]{0.22\linewidth}
        \includegraphics[width=1.\linewidth]{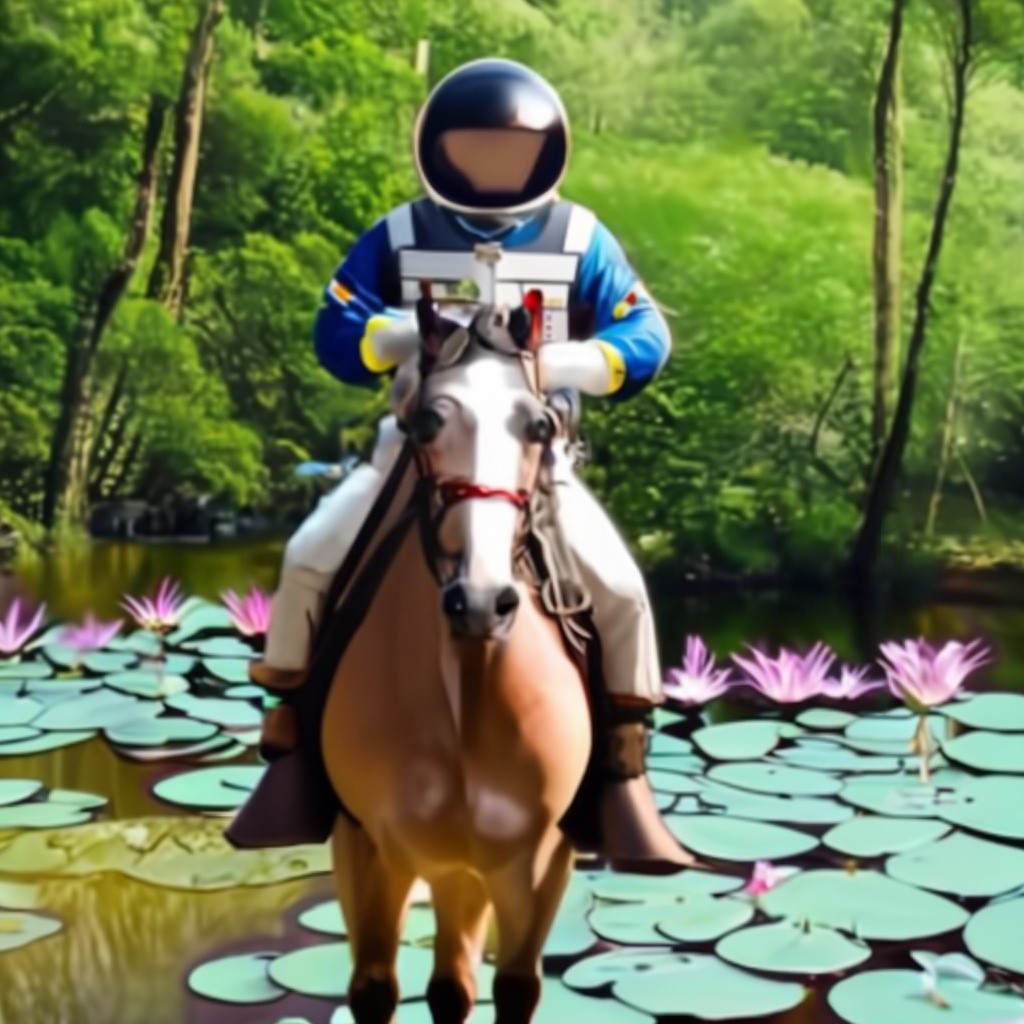}
        \caption{A photo of an astronaut riding a horse in the forest. There is a river in front of them with water lilies.}
    \end{subfigure}  ~
    \begin{subfigure}[t]{0.22\linewidth}
        \includegraphics[width=1.\linewidth]{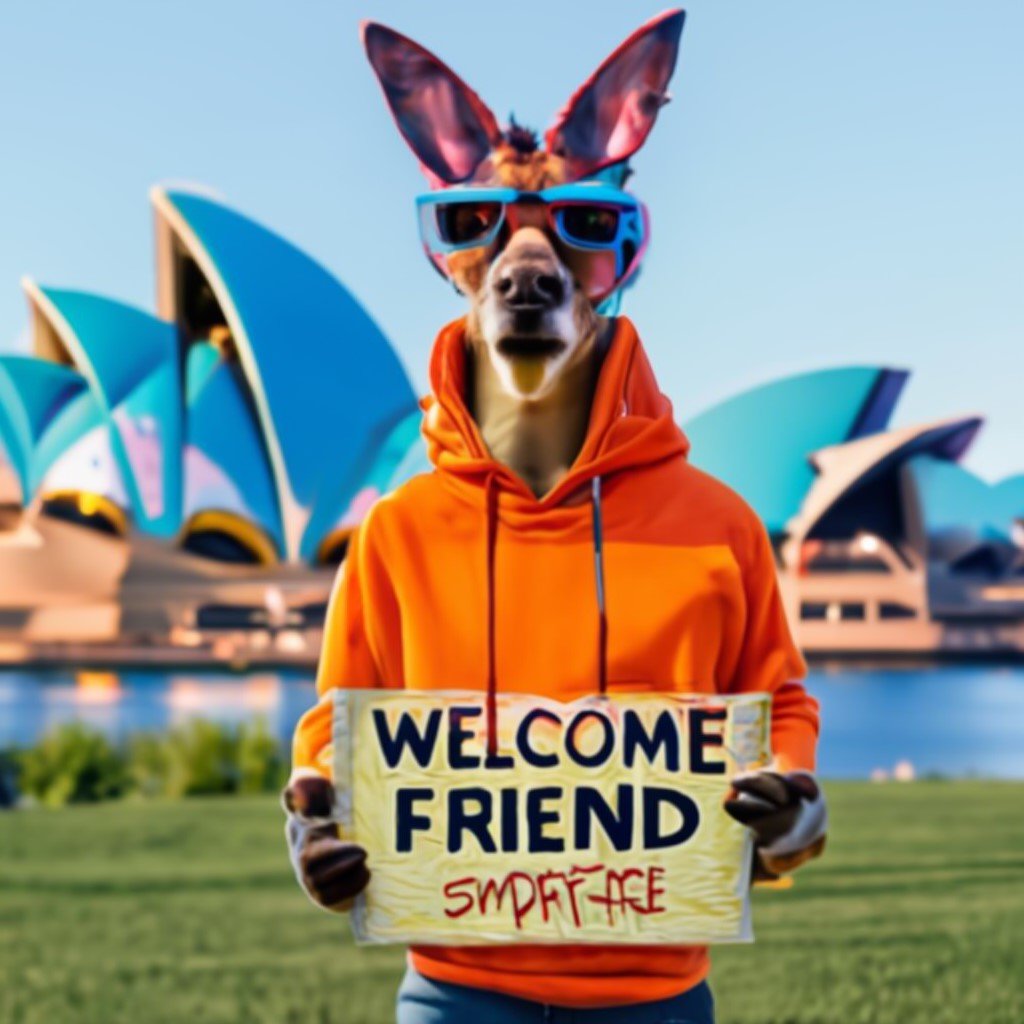}
        \caption{A portrait photo of a kangaroo wearing an orange hoodie and blue sunglasses standing on the grass in front of the Sydney Opera House holding a sign on the chest that says Welcome Friends!}
    \end{subfigure}     
    \vspace{-0.15in}
    \caption{Some generated examples on DrawBench and PartiPrompts.}
    \label{fig:generated_examples}
    \vspace{-0.3in}
\end{figure*}
\section{Experiments}

\subsection{Zero-shot Text-to-image Generation}
We first test the zero-shot text-to-image generation ability of Corgi. We prepare a dataset containing 900 million image-text pairs, which is composed of some commonly used datasets such as Conceptual Captions (CC3M)~\cite{Sharma2018ConceptualCA}, Conceptual Captions 12M (CC12M)~\cite{changpinyo2021conceptualcc12m}, filtered LAION-5B~\cite{schuhmann2021laion} and some image-text pairs collected by ourselves. Note that we make sure our dataset does not overlap with MS-COCO~\cite{lin2014microsoft}, CUB~\cite{WahCUB_200_2011}, Localized Narratives~\cite{pont2020connecting} and Multi-modal CelebA-HQ (MM-CelebA-HQ)~\cite{xia2021tedigan}, because we will test zero-shot or language-free performance on these downstream datasets. 
\paragraph{Decoder} Following \cite{nichol2021glide, ramesh2022hierarchical, saharia2022photorealistic}, our decoder adopts a model architecture consisting of a hierarchy of diffusion models, which has shown impressive ability in text-to-image generation. Specifically, three diffusion models are trained on 64, 256, and 1024 resolutions, respectively. All the 900M images are used to train the decoder. During training, each image is processed by three different pre-trained CLIP models: ViT-B/16, ViT-B/32, and RN-101. The outputs from these three models will be concatenated into a single 1536-dimensional embedding, which is then projected into eight vectors and fed into the decoder.

\paragraph{Prior model} To handle the image-text alignment, we train a shifted diffusion model to generate image embeddings from captions. Our shifted diffusion model is a decoder-only transformer whose input is a sequence consisting of encoded text from T5 \cite{2020t5}, CLIP text embedding, an embedding representing diffusion timestep, an embedding representing the index of corresponding Gaussian, a noised CLIP image embedding and a final embedding 
which will be used to predict the target CLIP image embedding. 
We train two variants on datasets with different scales: one prior is trained on the full 900M image-text pairs; the other one is trained on CC15M\footnote{collection of CC3M and CC12M}, which is a subset of our full dataset. 

\paragraph{Final model} Our two different prior models correspond to two final models. The decoders for the prior models (which are respectively trained on 900M and 15M image-text pairs) are trained on the 900M images. In other words, the two final models can be regarded as being trained in supervised and semi-supervised manners, where the training dataset of the first one consists of 900M image-text pairs, while the other one consists of 900M images but with only $1.7\%=15/900$ of them associating with captions. We denote these two variants as Corgi and Corgi-Semi, respectively. More implementation details are provided in the Appendix.

\begin{figure}[t!]
    \centering
    \begin{subfigure}[t]{0.45\linewidth}
        \includegraphics[width=1.\linewidth]{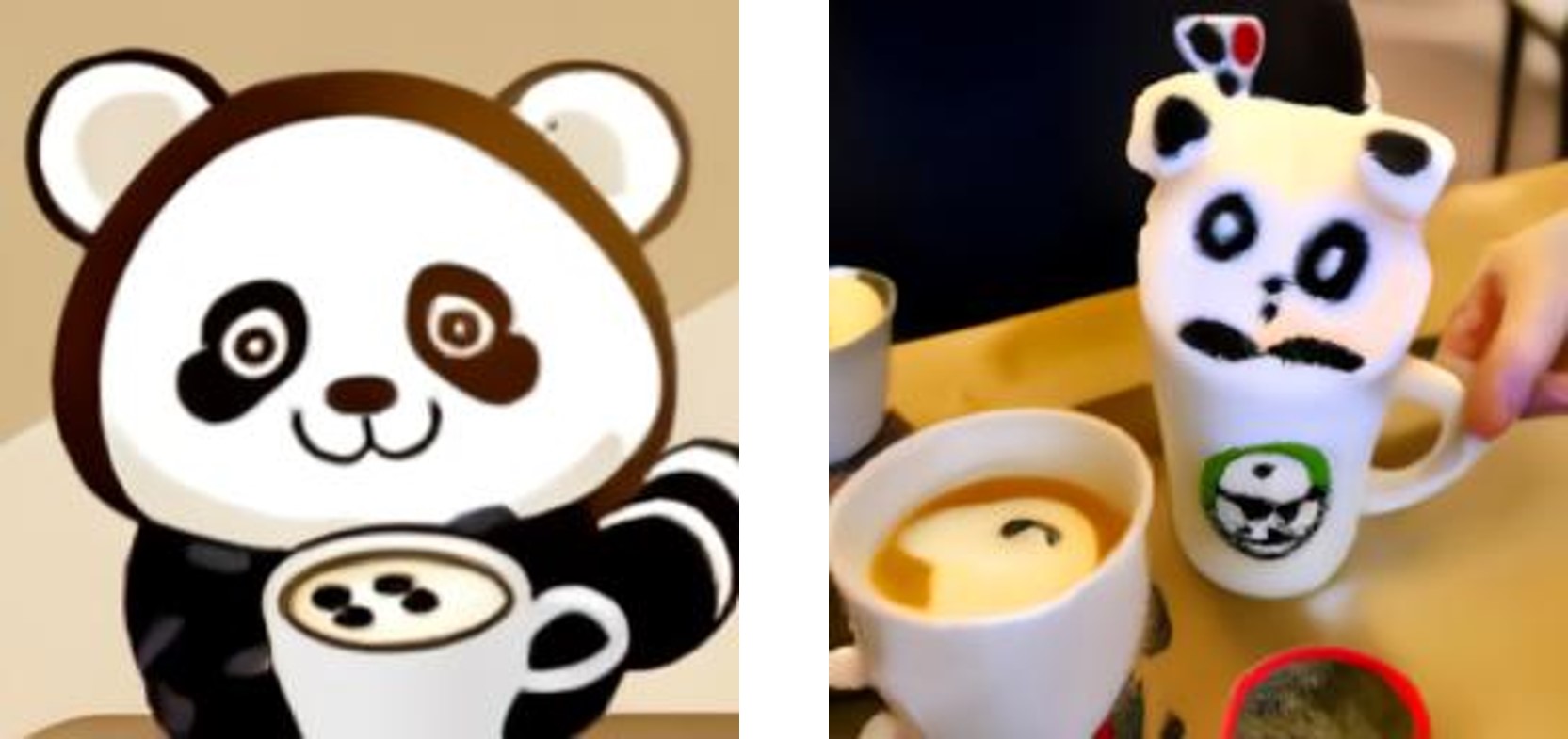}
        \caption{A panda making latte art.}
    \end{subfigure}
    \begin{subfigure}[t]{0.45\linewidth}
        \includegraphics[width=1.\linewidth]{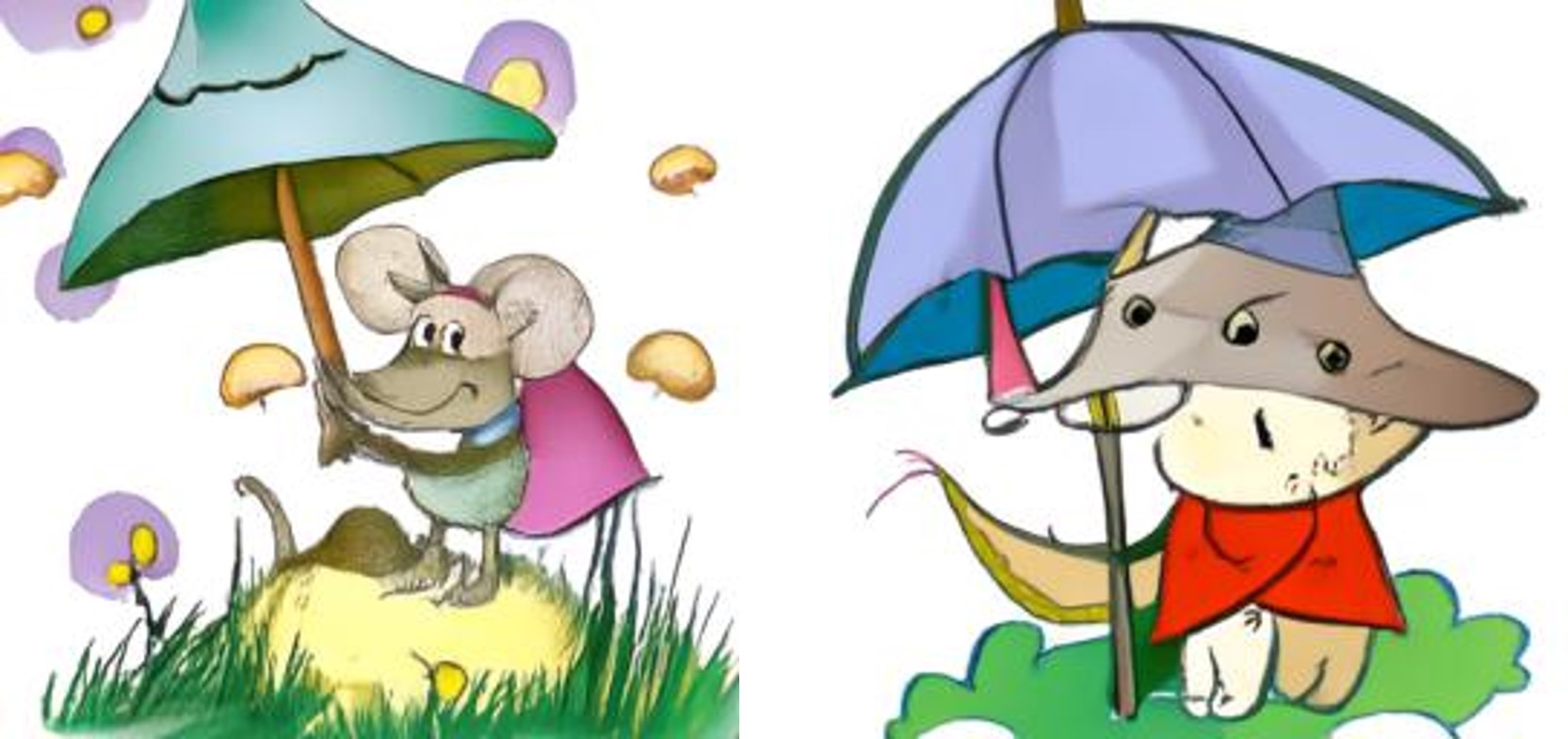}
        \caption{Illustration of a mouse using a mushroom as an umbrella.}
    \end{subfigure}       
    \vspace{-0.15in}
    \caption{Comparison of our final models trained in supervised (left) and semi-supervised manner (right).}
    \label{fig:dataset_comparison}
    \vspace{-0.15in}
\end{figure}

\begin{table}[t!]
\centering
\scalebox{0.75}{
    \begin{tabular}{lcc}
        \toprule
         Methods & Supervised FID $\downarrow$  & Zero-shot FID $\downarrow$ \\
         \midrule
         AttnGAN~\cite{xu2018attngan} & $33.10$ & \\
         DF-GAN~\cite{tao2021dfgan} & $21.42$ & \\
         XMC-GAN~\cite{zhang2021crossmodal} & $9.33$ & \\
         Lafite~\cite{zhou2021lafite} & $8.12$ & $26.94$ \\
        \midrule
        DALL-E~\cite{ramesh2021zero} & & $27.50$ \\
        CogView~\cite{ding2021cogview} & & $27.10$ \\
        GLIDE~\cite{nichol2021glide} & & $12.24$ \\
        LDM~\cite{rombach2021high} & & $12.63$\\
        Make-A-Scene~\cite{gafni2022make}& & $11.84$\\
        DALL-E 2~\cite{ramesh2022hierarchical} & & $10.39$\\
        CogView 2~\cite{ding2022cogview2} & & $24.00$ \\
        Imagen~\cite{saharia2022photorealistic} & & $7.27$ \\
        Parti~\cite{yu2022scaling} & & $7.23$ \\
        Re-Imagen~\cite{chen2022re} & & $6.88$ \\
        \midrule
        Corgi (Ours) & & $10.88$ \\
        Corgi-Semi (Ours) & & $10.60$ \\
         \bottomrule
    \end{tabular}
    }
    \vspace{-0.1in}
    \caption{Text-to-image generation results on MS-COCO.}    
    \label{tab:zero-shot}
    \vspace{-0.3in}
\end{table}



In Table \ref{tab:zero-shot}, we report zero-shot Fr\'echet Inception Distance (FID) evaluated on MS-COCO. Following previous works, the FID is calculated using 30,000 generated images, which corresponds to 30,000 randomly sampled captions from the validation set of MS-COCO. The results illustrate that our models obtain strong results, even when it is trained in a semi-supervised manner. Interestingly, our semi-supervised model obtains better FID than the supervised one. However, as we can see from Figure~\ref{fig:dataset_comparison}, the model trained with more image-text pairs leads to better image-text alignment and better image quality, as expected. One reason that it obtains relatively worse FID could be due to the dataset bias: CC15M might contain many samples that are similar to samples in MS-COCO, while this bias faded when they were merged to construct our full 900M dataset.

\begin{figure}[t!]
    \centering
    \includegraphics[width=0.85\linewidth]{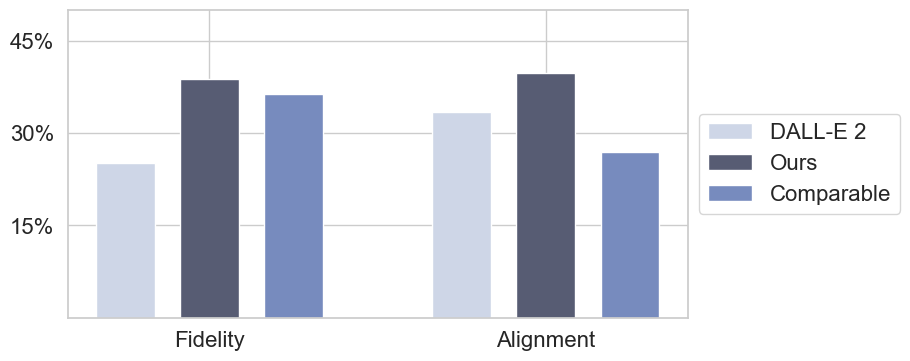}
    \vspace{-0.15in}
    \caption{Human evaluation on DrawBench and PartiPrompts.}
    \label{fig:human_evaluation}
    \vspace{-0.15in}
\end{figure}

We note that most large-scale text-to-image generation models are trained on different datasets, but all tested on MS-COCO, 
ignoring potential dataset bias. Comparing these models simply by FID might not be appropriate. Thus, comparing models by actual generation quality is necessary.
We follow previous works and choose to evaluate our model on DrawBench~\cite{saharia2022photorealistic} and PartiPrompts~\cite{yu2022scaling}. Some generated examples with complicated scenes are shown in Figure \ref{fig:generated_examples}.
We compare our model with DALL-E 2 and Stable Diffusion, as these are the only models allowing public access. For a fair comparison, we fine-tune our model for an extra 300,000 iterations so that the decoder can take both image embeddings and text as inputs, which is similar to DALL-E 2.
Some visual comparisons are provided in Figure \ref{fig:comparison_1}. More results are provided in the Appendix. As can be seen, our model leads to better generation in most cases in terms of image-text alignment and image fidelity. Furthermore, we conduct a human evaluation on DrawBench and PartiPrompts.
Specifically, we first generate four images for each prompt, then ten random human laborers are asked to judge which model is better (or comparable) in terms of image-text alignment and fidelity. The results are shown in Figure \ref{fig:human_evaluation}, where our model is shown to perform better in the two evaluation metrics consistently.

\begin{table}[t!]
\centering
\scalebox{0.75}{
    \begin{tabular}{lcccccc}
        \toprule
         & \multicolumn{2}{c}{MS-COCO}  & \multicolumn{2}{c}{CUB} &  \multicolumn{2}{c}{LN-COCO}\\
         Methods & IS $\uparrow$ & FID $\downarrow$ & IS $\uparrow$ & FID $\downarrow$ & IS $\uparrow$ & FID $\downarrow$ \\
         \midrule
         Lafite \cite{zhou2021lafite} & $27.20$ & $18.04$ & $4.32$ & $27.53$ & $18.49$ & $39.85$\\
         Lafite-2\cite{Zhou2022Lafite2FT} & $31.16$ &  $10.26$ & $4.93$ & $16.87$ & $23.18$ & $25.51$\\
         Corgi (Ours) & $\mathbf{34.14}$ & $10.33$ & $\mathbf{5.08}$ & $\mathbf{15.80}$ & $\mathbf{28.71}$ & $\mathbf{16.16}$\\
         \bottomrule
    \end{tabular}
    }
    \vspace{-0.1in}
    \caption{Language-free results on different datasets.}    
    \label{tab:language_free}
    \vspace{-0.3in}
\end{table}

\begin{figure*}[t!]
    \centering
    ~~~~~~
    \begin{subfigure}{0.75\linewidth}
        \centering
        \includegraphics[width=0.95\linewidth]{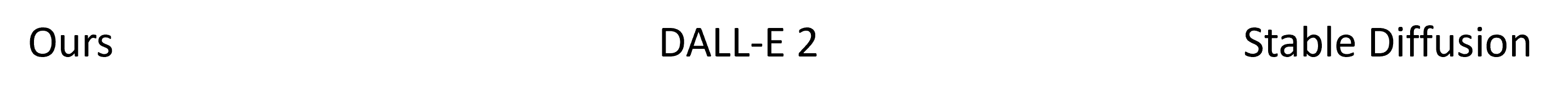}
    \end{subfigure}         
    \begin{subfigure}{0.95\linewidth}
    \centering
        \includegraphics[width=0.9\linewidth]{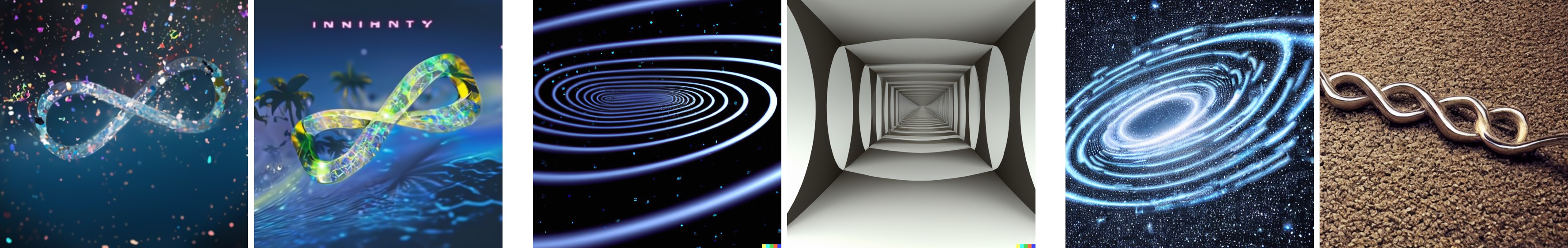}
        \caption{\textcolor{red}{Infinity}.}
    \end{subfigure}
    \begin{subfigure}{0.95\linewidth}
    \centering
        \includegraphics[width=0.9\linewidth]{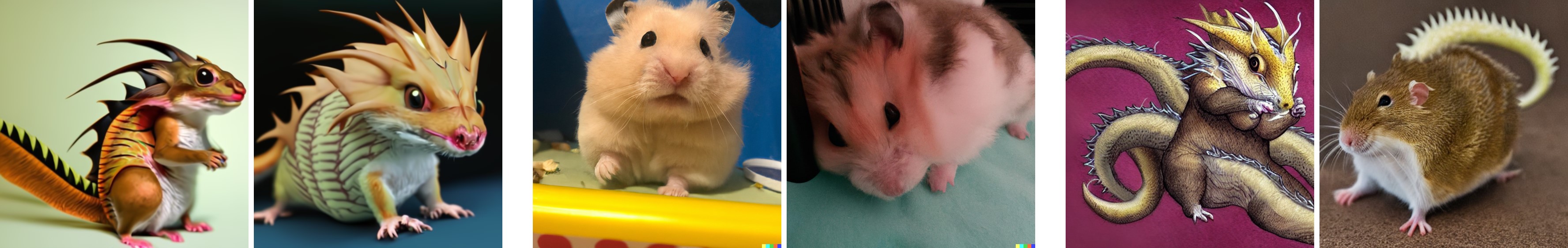}
        \caption{A \textcolor{red}{hamster dragon}.}
    \end{subfigure}    
    \begin{subfigure}{0.95\linewidth}
    \centering
        \includegraphics[width=0.9\linewidth]{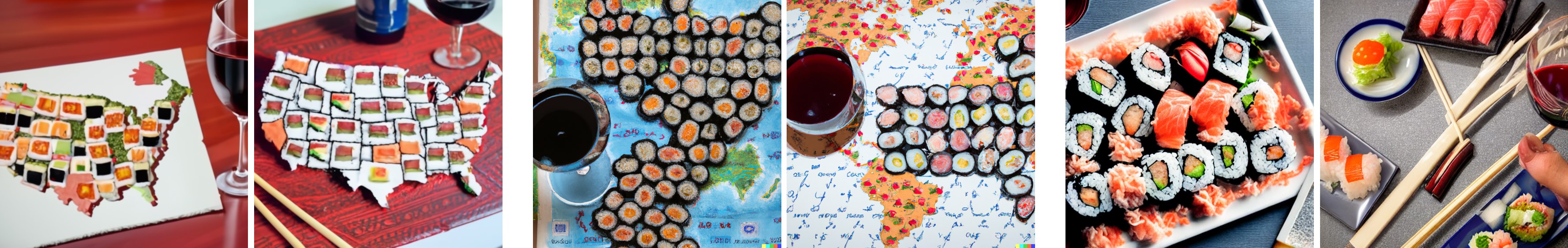}
        \caption{A  \textcolor{red}{map of the United States} made out \textcolor{red}{sushi}. It is on a table next to a \textcolor{red}{glass of red wine}.}
    \end{subfigure}     
    \begin{subfigure}{0.95\linewidth}
        \centering
        \includegraphics[width=0.9\linewidth]{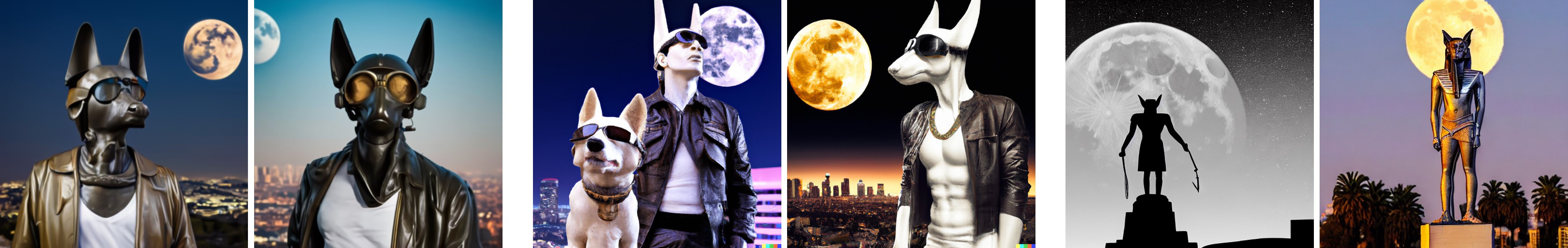}
        \caption{A portrait of a statue of the \textcolor{red}{Egyptian god Anubis} wearing \textcolor{red}{aviator goggles}, \textcolor{red}{white t-shirt} and \textcolor{red}{leather jacket}. A full \textcolor{red}{moon over the city} of Los Angeles is in the background at night.}
    \end{subfigure}    
    \begin{subfigure}{0.95\linewidth}
        \centering
        \includegraphics[width=0.9\linewidth]{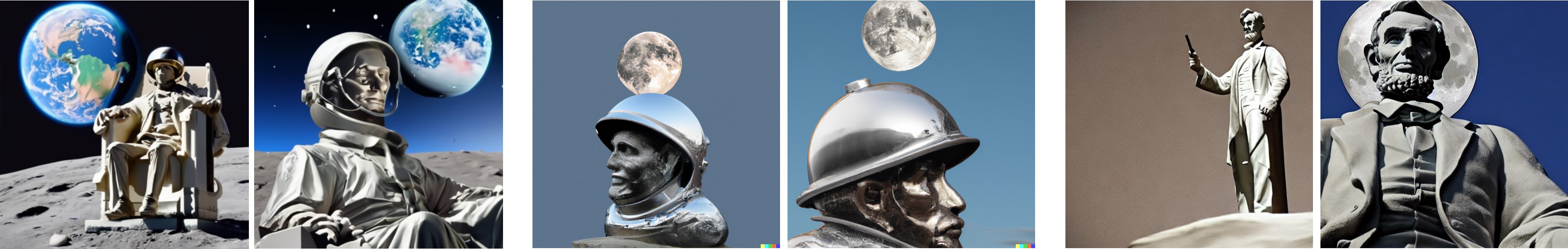}
        \caption{A statue of \textcolor{red}{Abraham Lincoln} wearing an opaque and shiny astronaut's \textcolor{red}{helmet}. The statue sits \textcolor{red}{on the moon}, with the planet \textcolor{red}{Earth in the sky}.}
    \end{subfigure}
    \vspace{-0.1in}
    \caption{Comparison with DALL-E 2 and Stable Diffusion. More results are provided in the Appendix.}
    \label{fig:comparison_1}
    \vspace{-0.3in}
\end{figure*}

\subsection{Language-free Text-to-image Generation}
\cite{zhou2021lafite} is the pioneering work to train text-to-image generation models on image-only datasets, termed language-free training as no associated captions are provided.
With a pre-trained shifted diffusion model, we can also perform language-free training and fine-tuning on any downstream dataset. Given an image-only dataset, we can train or fine-tune a generative model that generates images from image embeddings. Consequently, we can directly use the pre-trained shifted diffusion model at inference time to perform text-to-image generation.
Because our shifted diffusion model is pre-trained on large-scale image-text pairs, it is expected to generalize well on any downstream domain.

We first compare our method with \cite{zhou2021lafite, Zhou2022Lafite2FT}. For a fair comparison, our decoder uses the same network architecture as \cite{zhou2021lafite}, which is a StyleGAN2-based model.
Some quantitative results are provided in Table \ref{tab:language_free}, from which we can see that our method leads to better results in general. 
More results are provided in the Appendix. 
Although Lafite-2 \cite{Zhou2022Lafite2FT} achieves a competitive FID on the MS-COCO dataset, it was trained on pseudo captions that require extra human workload. Specifically, to train Lafite-2, domain-specific vocabularies and prompts are needed, which requires human prior knowledge for each downstream domain. On the other hand, our shifted diffusion model is pre-trained and can be directly plugged into any domain without further training or fine-tuning.

\begin{figure}[t!]
    \centering
    \includegraphics[width=0.85\linewidth]{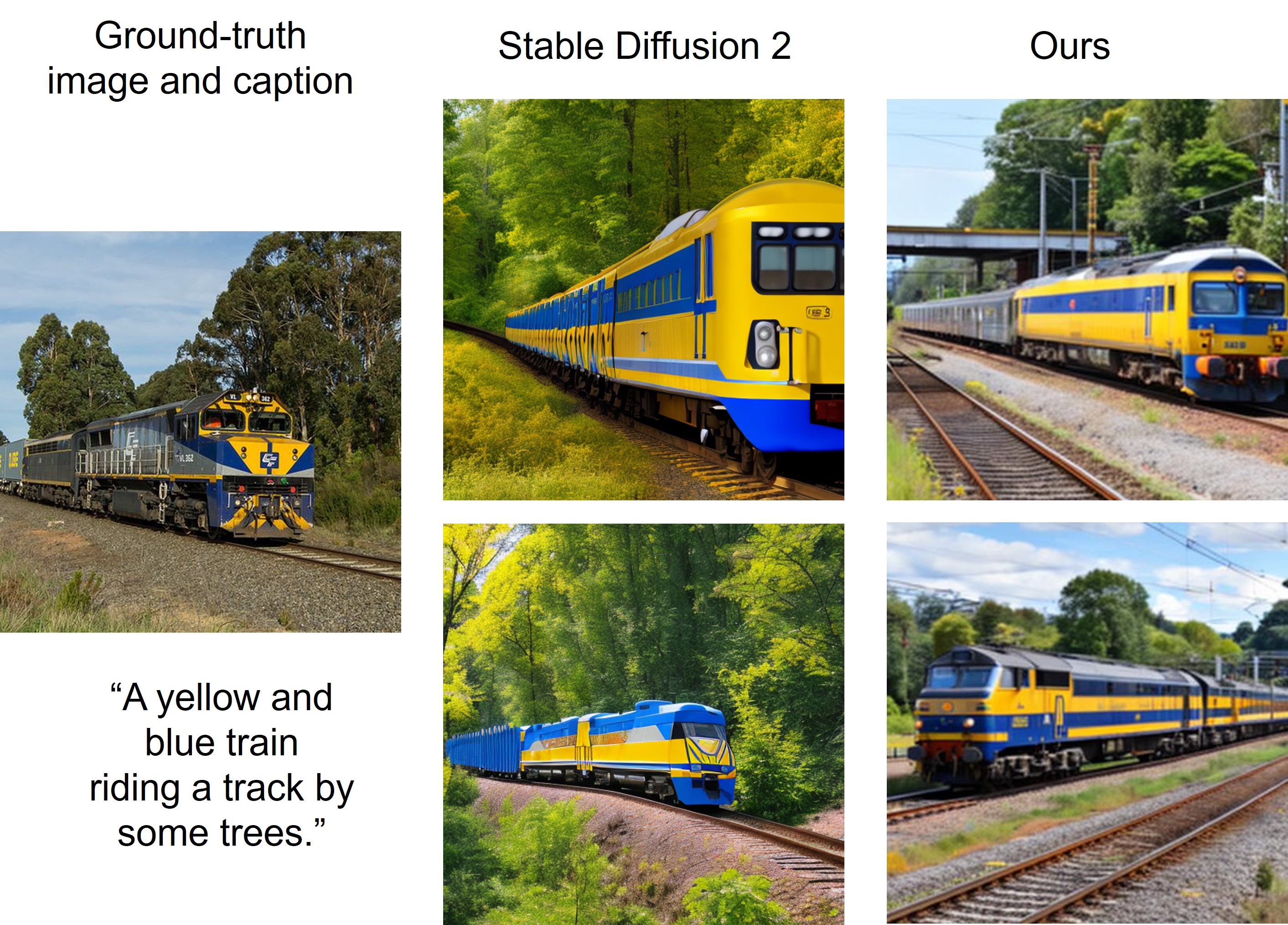}
    \vspace{-0.1in}
    \caption{Pre-trained Stable Diffusion 2 vs. our fine-tuned model.}
    \label{fig:pre-training_fine-tuning}
    \vspace{-0.15in}
\end{figure}

We then conduct experiments of fine-tuning pre-trained text-to-image diffusion model under language-free setting. We choose Stable Diffusion 2 as our base model. Specifically, a projection layer is added to project CLIP image embedding into 4 vectors, which will be fed into the UNet of Stable Diffusion 2. As shown in Figure \ref{fig:pre-training_fine-tuning}, Stable Diffusion 2 may generate images whose styles are different from target dataset, while our model leads to more satisfactory results after language-free fine-tuning.
\subsection{Ablation Study}

\paragraph{Baseline vs. shifted diffusion}
Although we have shown that our model obtains better generation quality than DALL-E 2 in previous experiments, there is no direct clue suggesting that the improvement is due to the advantages of shifted diffusion over the baseline diffusion. This is because, as we pointed out, other important factors, such as the potential dataset bias and different implementation details, may also affect model performance. 
Therefore, to better compare our shifted diffusion model with the baseline diffusion model, we conduct an ablation study where we train these two diffusion priors with the same implementation and training dataset (CC15M). 
We set $k=1$ for shifted diffusion for a fair comparison, as a larger $k$ leads to further improvements. More details are provided in the Appendix.

\begin{figure}[t!]
    \centering
    \begin{subfigure}[b]{0.350\linewidth}
        \includegraphics[width=1.\linewidth]{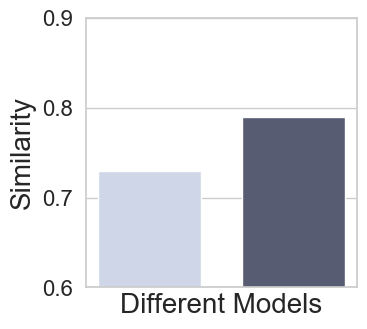}
    \end{subfigure}    
    \begin{subfigure}[b]{0.55\linewidth}
        \includegraphics[width=1.\linewidth]{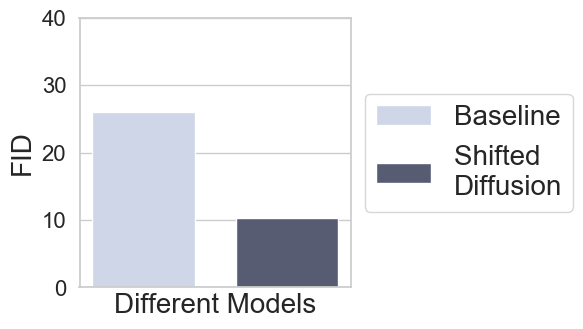}
    \end{subfigure}   
    \vspace{-0.15in}
    \caption{Comparison between baseline and shifted diffusion on our diffusion-based decoder which is trained from scratch.}
    \label{fig:ablation_1}
    \vspace{-0.25in}
\end{figure}

Firstly, we compare the cosine similarity of the generated image embeddings to the ground-truth image embeddings. We randomly sample 10,000 image-text pairs from the validation set of MS-COCO to prevent overlapping between training and testing datasets. The sampled captions are fed into different models to generate corresponding image embeddings. 
The results are shown in Figure \ref{fig:ablation_1}. We can see that our shifted diffusion model leads to higher similarity scores than the baseline, implying better embedding generation. 

\begin{figure}[t!]
    \centering
    \includegraphics[width=0.5\linewidth]{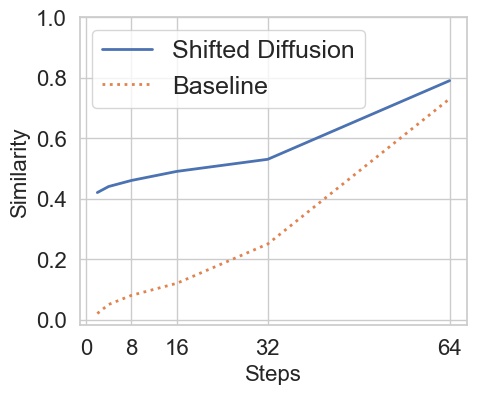}
    \vspace{-0.15in}
    \caption{Evolution of embedding similarity during sampling.}
    \label{fig:baseline_sft_steps}
    \vspace{-0.18in}
\end{figure}

Next, we compare the similarity scores of the generated embeddings to ground truth at different timesteps during the sampling process. We use 64 strided steps for both baseline and shifted diffusion model. The results are shown in Figure \ref{fig:baseline_sft_steps}, from which we can see that our method leads to higher similarity scores, especially at initialization. This indicates that our starting point is much closer to the target, consistent with our design intention.

\begin{figure}[t!]
    \centering
    \begin{subfigure}[t]{0.45\linewidth}
        \includegraphics[width=1.\linewidth]{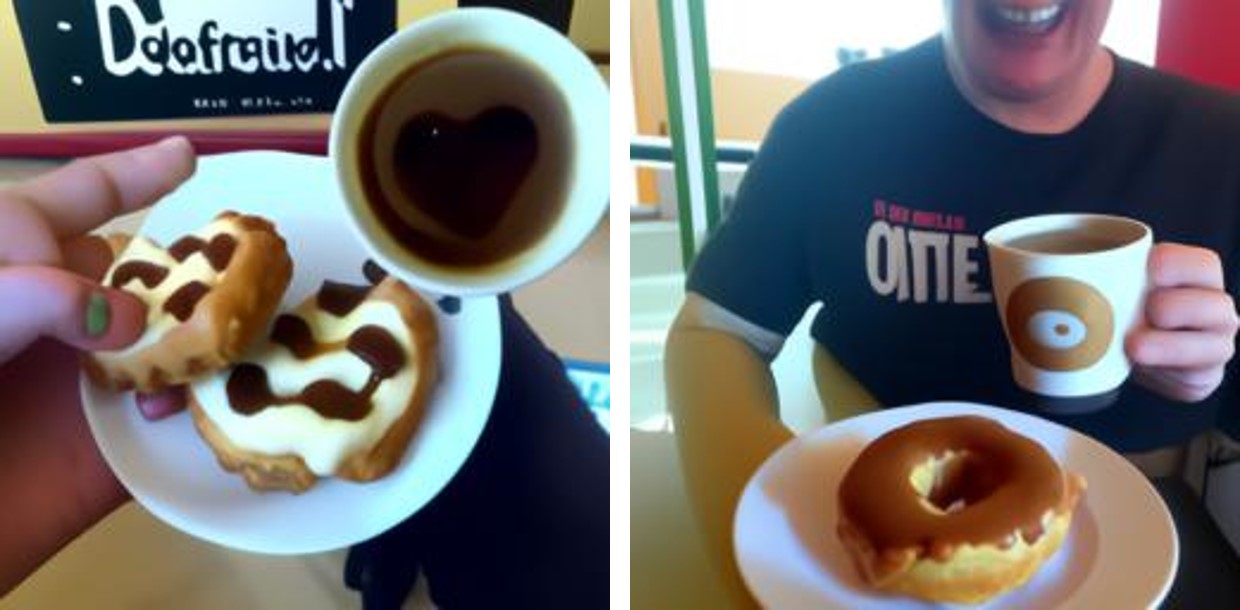}
        \centering
        \caption{A close-up photo of a person holding a cup and a doughnut}
    \end{subfigure}   ~
    \begin{subfigure}[t]{0.45\linewidth}
        \includegraphics[width=1.\linewidth]{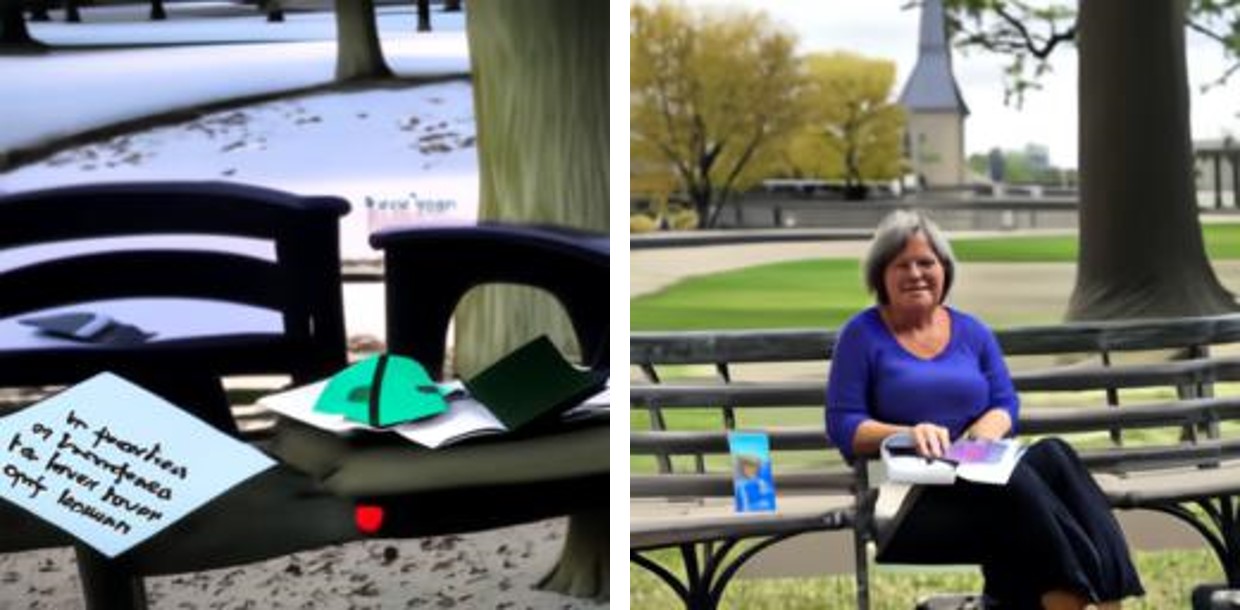}
        \centering
        \caption{A lady sitting on a park bench reading a book.}
    \end{subfigure}    
    \vspace{-0.1in}
    \caption{Generated examples with baseline diffusion (left) and shifted diffusion (right).}
    \vspace{-0.2in}
    \label{fig:ablation_2}
\end{figure}

\begin{figure}[t!]
    \centering
    \begin{subfigure}[t]{0.480\linewidth}
        \includegraphics[width=1.\linewidth]{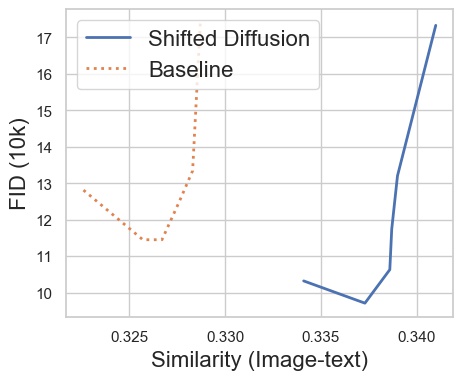}
    \end{subfigure}   
    \begin{subfigure}[t]{0.480\linewidth}
        \includegraphics[width=1.\linewidth]{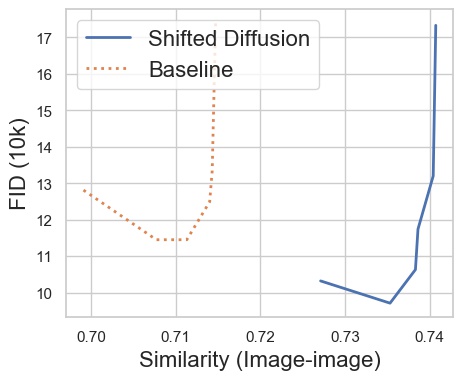}
    \end{subfigure}   
    \vspace{-0.1in}
    \caption{Comparison between baseline and shifted diffusion on fine-tuned Stable Diffusion 2.}
    \label{fig:fid_clip}
    \vspace{-0.2in}
\end{figure}

Finally, we compare the shifted diffusion and baseline by applying them in text-to-image generation tasks. 
To this end, we first generate 30,000 images using our diffusion-based decoder which is trained from scratch. The zero-shot FID results on MS-COCO are shown in Figure \ref{fig:ablation_1}, along with some generated examples in Figure \ref{fig:ablation_2}. We can see that shifted diffusion indeed leads to better quantitative and qualitative results, while the baseline model fails to capture some details in the text. We also evaluate them on fine-tuned Stable Diffusion 2, where FID and CLIP similarities of 10,000 generated images to ground-truth images and captions are calculated. The results are provided in Figure \ref{fig:fid_clip}\footnote{We report average similarity evaluated by ViT-B/16, ViT-B/32 and RN-101 CLIP models.}, from which we can find that shifted diffusion leads to better results as it obtains lower FID and higher CLIP similarities.

\begin{figure}[t!]
    ~~~~~~~~~~~~~~~~~~~~~
    \includegraphics[width=0.7\linewidth]{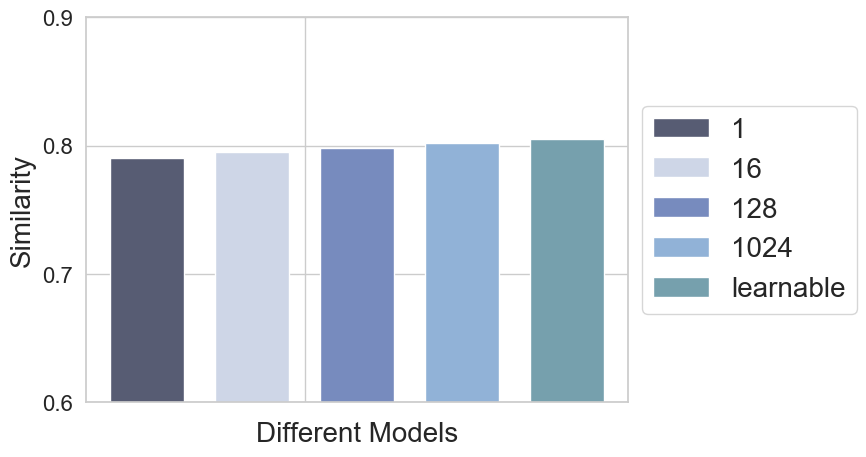}
    \vspace{-0.15in}
    \caption{Results of shifted diffusion with different settings.}
    \label{fig:sft_similarities}
    \vspace{-0.25in}
\end{figure}

\paragraph{Different settings for shifted diffusion}
Recall that our shifted diffusion model adopts one or more Gaussian distributions as its initialization for the sampling process. We investigate the influence of the number of distributions in this experiment. We train our shifted diffusion model with 1, 16, 128, and 1024 Gaussians, respectively. Furthermore, we train a model with 1024 learn-able mean vectors and covariance matrices. All the models are trained on CC15M. We calculate the similarity of generated image embeddings to the ground-truth embeddings on the validation set of MS-COCO. The results are shown in Figure \ref{fig:sft_similarities}, from which we can see that using more Gaussian distributions leads to better results; furthermore, making the parameters learn-able gains further performance improvement. Some more discussions are provided in the Appendix.

\section{Conclusion}
We propose Corgi, a novel and general diffusion model that benefits text-to-image generation under different settings. Extensive large-scale experiments are conducted. Strong quantitative and qualitative results are obtained, illustrating the effectiveness of the proposed method. 

\clearpage
{\small
\bibliographystyle{ieee_fullname}
\bibliography{egbib}
}

\clearpage
\appendix
\onecolumn
\section{Derivation of Shifted Diffusion}
Recall that we define 
\[
    q(\zb_t \vert \zb_{t-1}) \coloneqq \mathcal{N}(\zb_t; \sqrt{1 - \beta_t} \zb_{t-1} + \sbb_t, \beta_t \Sigmab),
\]
we use deduction to prove that 
\begin{align}\label{eq:proof_closed_form}
    q(\zb_t \vert \zb_0) = \mathcal{N}(\zb_t; \sqrt{\bar{\alpha}_{t}} \zb_0 +  \sum_{i=1}^{t} \sbb_i\sqrt{\bar{\alpha}_{t}/\bar{\alpha}_i}, (1 - \bar{\alpha}_{t})\Sigmab).
\end{align}

When $t=1$, we know that 
\[
 q(\zb_1 \vert \zb_{0}) =  \mathcal{N}(\zb_1; \sqrt{1 - \beta_1} \zb_{0} + \sbb_1, \beta_1 \Sigmab).
\]
We can re-write
\begin{align*}
    \sqrt{1 - \beta_1} \zb_{0} + \sbb_1 &=  \sqrt{1 - \beta_1} \zb_{0} + \sum_{i=1}^1 \sbb_i \sqrt{\bar{\alpha}_1/\bar{\alpha}_i}\\
    &= \sqrt{\bar{\alpha}_1} \zb_{0}  +  \sum_{i=1}^1 \sbb_i  \sqrt{\bar{\alpha}_1/\bar{\alpha}_i}
\end{align*}
and 
\[
\beta_1 \Sigmab = \{1 - (1 - \beta_1)\}\Sigmab  = (1 - \bar{\alpha}_1) \Sigmab,
\]
which means \eqref{eq:proof_closed_form} holds when $t=1$.

Now we prove the case when $t>1$. Assume Equation \ref{eq:proof_closed_form} holds for time $t=s$, which means
\[
    q(\zb_s \vert \zb_0) = \mathcal{N}(\zb_s; \sqrt{\bar{\alpha}_{s}} \zb_0 +  \sum_{i=1}^{s} \sbb_i\sqrt{\bar{\alpha}_{s}/\bar{\alpha}_i}, (1 - \bar{\alpha}_{s})\Sigmab).
\]
We also know that
\[
    q(\zb_{s+1} \vert \zb_{s}) \coloneqq \mathcal{N}(\zb_{s+1}; \sqrt{1 - \beta_s} \zb_{s} + \sbb_{s+1}, \beta_s \Sigmab),
\]
by the property of Gaussian distribution~\cite{bishop2006PRML}, we know that $q(\zb_{s+1} \vert \zb_0)$ is also a Gaussian distribution with mean 
\begin{align*}
    &(\sqrt{\bar{\alpha}_s}\zb_0 + \sum_{i=1}^{s} \sbb_i\sqrt{\bar{\alpha}_{s}/\bar{\alpha}_i})\sqrt{1 - \beta_{s+1}} + \sbb_{s+1}\\
    = & (\sqrt{\bar{\alpha}_s}\zb_0 + \sum_{i=1}^{s} \sbb_i\sqrt{\bar{\alpha}_{s}/\bar{\alpha}_i})\sqrt{1 - \beta_{s+1}} + \sbb_{s+1}\sqrt{\bar{\alpha}_{t+1} / \bar{\alpha}_{t+1}} \\
    = & \sqrt{\bar{\alpha}_{s+1}}\zb_0 + \sum_{i=1}^{s} \sbb_i\sqrt{\bar{\alpha}_{s+1}/\bar{\alpha}_i} + \sbb_{s+1}\sqrt{\bar{\alpha}_{t+1} / \bar{\alpha}_{t+1}} \\
    = & \sqrt{\bar{\alpha}_{s+1}}\zb_0 + \sum_{i=1}^{s+1} \sbb_i\sqrt{\bar{\alpha}_{s+1}/\bar{\alpha}_i},
\end{align*}
and covariance matrix (a diagonal matrix)
\begin{align*}
    &\beta_{s+1} \Sigmab + \sqrt{1 - \beta_{s+1}}(1 - \bar{\alpha}_s) \Sigmab \sqrt{1 - \beta_{s+1}}\\
    = & \beta_{s+1} \Sigmab + (1 - \beta_{s+1})(1 - \bar{\alpha}_s) \Sigmab \\
    = & \Sigmab \{ \beta_{s+1} + (1 - \beta_{s+1})(1 - \bar{\alpha}_s)\} \\
    = & \Sigmab (\beta_{s+1} + 1 - \beta_{s+1} - \bar{\alpha}_s + \beta_{s+1}\bar{\alpha}_s) \\
    = & \Sigmab (1 - \bar{\alpha}_s + \beta_{s+1}\bar{\alpha}_s) \\
    = & \Sigmab \{ 1 - (1 - \beta_{s+1})\bar{\alpha}_s\}\\
    = & \Sigmab (1 - \bar{\alpha}_{s+1}),
\end{align*}
which means Equation \ref{eq:proof_closed_form} holds for $t = s+1$ when it holds for $t = s$. By deduction, we know that since it holds for $t=1$, it holds for any integer. 

Specifically, if we choose $\sbb_t = (1 - \sqrt{1 - \beta_t})\mub$, we get
\[
    q(\zb_t \vert \zb_0) = \mathcal{N}(\zb_t; \sqrt{\bar{\alpha}_{t}} \zb_0 + (1 - \sqrt{\bar{\alpha}_{t}})\mub, (1 - \bar{\alpha}_{t})\Sigmab),
\]
because
\begin{align*}
    & \sqrt{\bar{\alpha}_{t}}\zb_0 + \sum_{i=1}^{t} \sbb_i\sqrt{\dfrac{\bar{\alpha}_{t}}{\bar{\alpha}_i}} \\
    = & \sqrt{\bar{\alpha}_{t}}\zb_0 + \sum_{i=1}^{t}  (1 - \sqrt{1 - \beta_i})\mub\sqrt{\dfrac{\bar{\alpha}_{t}}{\bar{\alpha}_i}}\\
    = & \sqrt{\bar{\alpha}_{t}}\zb_0 + \mub \sum_{i=1}^{t}  (1 - \sqrt{1 - \beta_i})\sqrt{\dfrac{\bar{\alpha}_{t}}{\bar{\alpha}_i}}\\
    = &  \sqrt{\bar{\alpha}_{t}}\zb_0 + \mub (\sqrt{\dfrac{\bar{\alpha}_{t}}{\bar{\alpha}_{t}}} - \sqrt{\dfrac{\bar{\alpha}_{t}}{\bar{\alpha}_{t-1}}} + \sqrt{\dfrac{\bar{\alpha}_{t}}{\bar{\alpha}_{t-1}}} - \sqrt{\dfrac{\bar{\alpha}_{t}}{\bar{\alpha}_{t-2}}} +...+\sqrt{\dfrac{\bar{\alpha}_{t}}{\bar{\alpha}_{2}}} - \sqrt{\dfrac{\bar{\alpha}_{t}}{\bar{\alpha}_{1}}} + \sqrt{\dfrac{\bar{\alpha}_{t}}{\bar{\alpha}_{1}}} - \sqrt{\bar{\alpha}_t}) \\ 
    = & \sqrt{\bar{\alpha}_{t}} \zb_0 + (1 - \sqrt{\bar{\alpha}_{t}})\mub.
\end{align*}

We are now able to get the closed-form expression of posterior $q(\zb_{t-1} \vert \zb_t, \zb_0)$. We know that
\begin{align*}
    q(\zb_{t} \vert \zb_0) &= \mathcal{N}(\zb_t; \sqrt{\bar{\alpha}_t} \zb_0 + (1 - \sqrt{\bar{\alpha}_t})\mub, (1 - \bar{\alpha}_t)\Sigmab) \\
    q(\zb_{t-1} \vert \zb_0) &= \mathcal{N}(\zb_{t-1}; \sqrt{\bar{\alpha}_{t-1}} \zb_0 + (1 - \sqrt{\bar{\alpha}_{t-1}})\mub, (1 - \bar{\alpha}_{t-1})\Sigmab) \\
    q(\zb_t \vert \zb_{t-1}) &= \mathcal{N}(\zb_t; \sqrt{1 - \beta_t} \zb_{t-1} + (1 - \sqrt{1 - \beta_t})\mub, \beta_t \Sigmab).
\end{align*}
From \cite{bishop2006PRML}, we know that 
\begin{align*}
    q(\zb_{t-1}\vert \zb_t, \zb_0) = \mathcal{N}(\zb_{t-1}; \nub, \mathbf{\Lambda})
\end{align*}
where
\begin{align*}
    \nub &= \mathbf{\Lambda}\{\sqrt{1 - \beta_t}\{\zb_t - (1 - \sqrt{1 - \beta_t})\mub\}\Sigmab^{-1}/\beta_t + \{\sqrt{\bar{\alpha}_{t-1}} \zb_0 + (1 - \sqrt{\bar{\alpha}_{t-1}})\mub\} \Sigmab^{-1}/(1 - \bar{\alpha}_{t-1})\},\\
     \mathbf{\Lambda} &=\{ \Sigmab^{-1}/(1 - \bar{\alpha}_{t-1}) + (1 - \beta_t)\Sigmab^{-1}/\beta_t\}^{-1}.
\end{align*}
Equation \eqref{eq:q_posterior} can be obtained by simple derivation.

\section{More Experimental Results}
\paragraph{Language-free text-to-image generation}
We provide more results on language-free text-to-image generation in Table \ref{tab:language_free_2}. Results in Table \ref{tab:language_free_2} are models trained from scratch on corresponding datasets. Results of MM-CelebA-HQ in \cite{Zhou2022Lafite2FT} is based on a model which is first pre-trained on FFHQ dataset~\cite{karras2019style}. For fair comparison, we use the code provided by the authors and train Lafite-2 from scratch on MM-CelebA-HQ, without pre-training on FFHQ.
\begin{table}[h!]
\centering
\scalebox{0.85}{
    \begin{tabular}{lcccccccc}
        \toprule
         & \multicolumn{2}{c}{MS-COCO}  & \multicolumn{2}{c}{CUB} &  \multicolumn{2}{c}{LN-COCO}&  \multicolumn{2}{c}{MM-CelebA-HQ}\\
         Methods & IS $\uparrow$ & FID $\downarrow$ & IS $\uparrow$ & FID $\downarrow$ & IS $\uparrow$ & FID $\downarrow$ & IS $\uparrow$ & FID $\downarrow$ \\
         \midrule
         Lafite \cite{zhou2021lafite} & $27.20$ & $18.04$ & $4.32$ & $27.53$ & $18.49$ & $39.85$ & $2.89$ & $32.75$\\
         Lafite-2\cite{Zhou2022Lafite2FT} & $31.16$ &  $10.26$ & $4.93$ & $16.87$ & $23.18$ & $25.51$ & $2.91$ & $21.89$\\
         Corgi (Ours) & $\mathbf{34.14}$ & $10.33$ & $\mathbf{5.08}$ & $\mathbf{15.80}$ & $\mathbf{28.71}$ & $\mathbf{16.16}$ & $\mathbf{3.06}$ & $\mathbf{19.74}$\\
         \bottomrule
    \end{tabular}
    }
    \caption{Language-free results on MS-COCO, CUB, LN-COCO and MM-CelebA-HQ datasets.}    
    \label{tab:language_free_2}
\end{table}

\paragraph{Ablation study on shifted diffusion}
\begin{wrapfigure}{r}{0.5\textwidth}
    \centering
    \begin{subfigure}[t]{0.45\linewidth}
        \includegraphics[width=1.\linewidth]{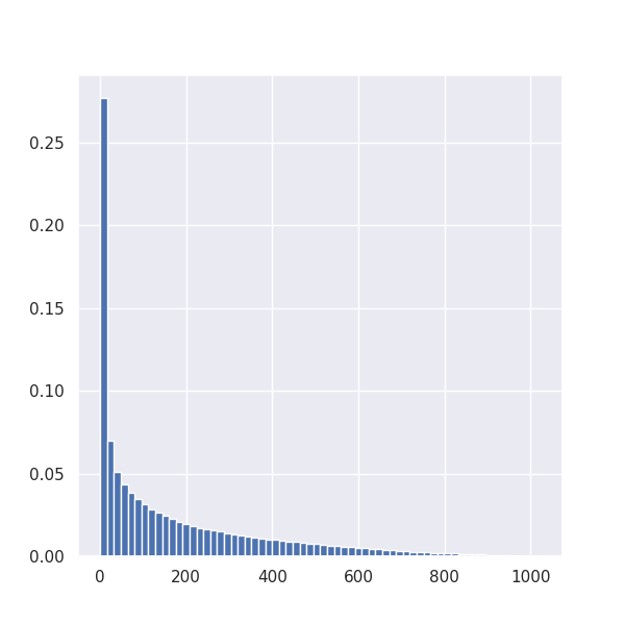}
        \caption{1024 fixed Gaussian}
    \end{subfigure}     
    \begin{subfigure}[t]{0.45\linewidth}
        \includegraphics[width=1.\linewidth]{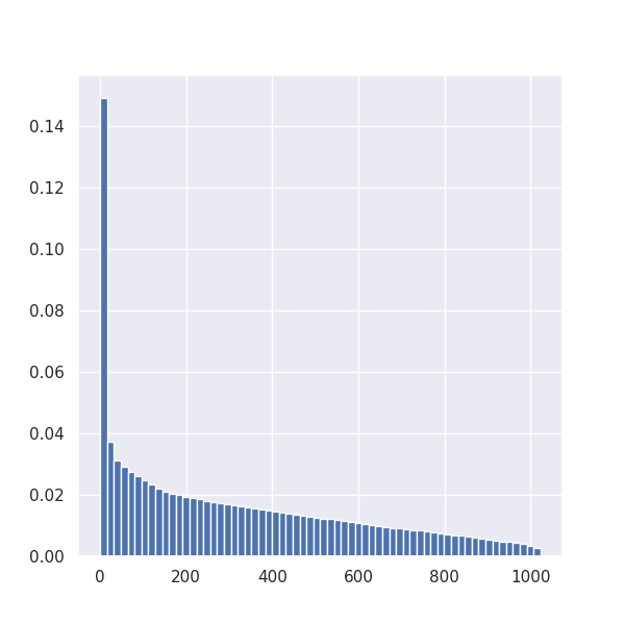}
        \caption{1024 learn-able Gaussian}
    \end{subfigure}         
    \caption{Frequency of clusters being selected.}
    \label{fig:frequency}
\end{wrapfigure}
We already know that learn-able Gaussians may lead to higher similarity of generated image embedding to ground-truth embedding, we would like to further compare them. We train 2 shifted diffusion models, with 1024 fixed/learn-able Gaussian distributions. After training, we feed random text captions from the validation set of MS-COCO to the model, and calculate the frequency of each Gaussian being selected by \eqref{eq:assign}. We rank and plot the frequency in Figure \ref{fig:frequency}. Intuitively, when we set $k=1024$, we are expecting 1024 representative clusters with different semantics, thus we hope every Gaussian has a chance of being selected. 
However, as we can see in the figure, in the case where 1024 fixed Gaussians are introduced, some Gaussians are never selected which means they contribute nothing to the model. On the contrary, with learned mean and covariance matrix, every Gaussian has a chance of being selected.

\begin{figure}
    \centering
    \includegraphics[width=0.95\linewidth]{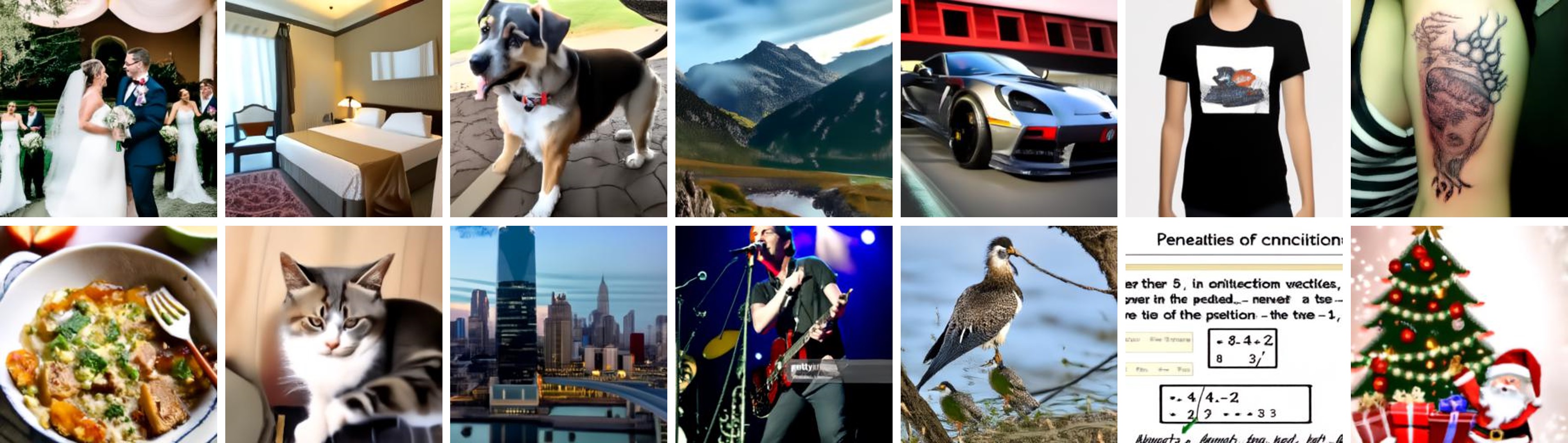}
    \caption{Generated images of mean embeddings from different Gaussian clusters.}
    \label{fig:means}
\end{figure}

A natural question is, are the clusters semantically different? To answer this question, we generate images by directly feeding means of different clusters, which are shown in Figure \ref{fig:means}. As we can see, embeddings corresponding to different clusters will lead to generated images that have very different semantics.

\section{Implementation Details}
Our prior model trained on 900M dataset is a decoder-only transformer. We set the width, depth, number of attention heads to be 2048, 20, 32 respectively. The model is trained for 500,000 iterations with a batch size of 4096. For the prior model trained on CC15M and baseline prior model in ablation study, we reduce the transformer depth from 20 to 16, while keeping the width and number of attention heads unchanged. The smaller prior models are trained for 40,000 iterations with batch size of 4096 on CC15M dataset. AdamW~\cite{loshchilov2018decoupledAdamW} optimizer with learning rate of $1.2\times 10^{-4}$, $\beta=(0.9, 0.96)$, $\epsilon=10^{-6}$. We drop the encoded text with probability of 0.1 to enable classifier-free guidance sampling.~\cite{ho2021classifier}

Our diffusion-based decoder follows DALL-E 2~\cite{ramesh2022hierarchical}. Batch sizes are set to be 2048, 1024, 512 for diffusion models at 64, 256, 1024 resolutions respectively. The models are trained for 1,000,000 iterations. AdamW optimizer with $\beta=(0.9, 0.999), \epsilon=10^{-8}$ is used for all three models. Learning rate is set to be $10^{-4}$ for the model at 1024 resolution, while it is set to be $1.2\times 10^{-4}$ for the other two models.  We drop the encoded text with probability of 0.1 to enable classifier-free guidance sampling.

Our GAN-based decoder follows the design in \cite{zhou2021lafite}. The batch size is set to be 64. Adam optimizer~\cite{kingma2014adam} with learning rate 0.0025, $\beta=(0, 0.99), \epsilon=10^{-8}$ is used for both generator and discriminator. 

\section{More Generated Examples}

We provide more generated examples and comparisons here.
\begin{figure*}[ht!]
    \centering
    ~~~~~~~~
    \begin{subfigure}{0.8\linewidth}
        \includegraphics[width=1.\linewidth]{figures/Comparison/title.png}
    \end{subfigure}         
    \begin{subfigure}{0.95\linewidth}
        \includegraphics[width=1.\linewidth]{figures/Comparison/c_infinity.jpg}
        \caption{Infinity.}
    \end{subfigure}
    \begin{subfigure}{0.95\linewidth}
        \includegraphics[width=1.\linewidth]{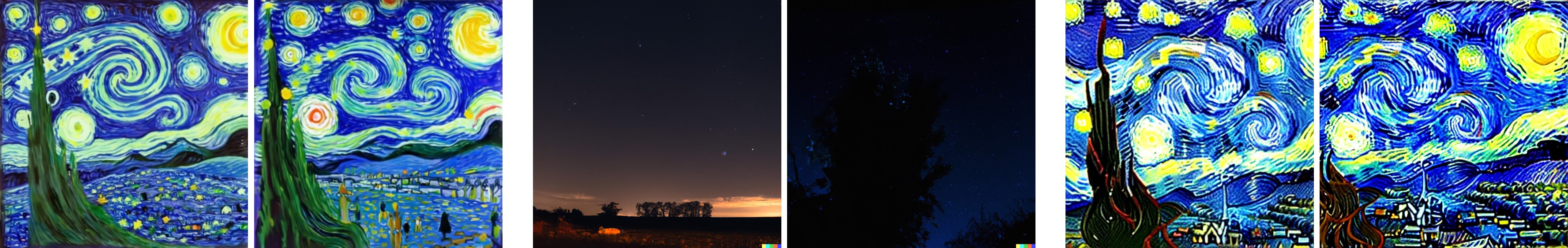}
        \caption{The Starry Night.}
    \end{subfigure}    
    \begin{subfigure}{0.95\linewidth}
        \includegraphics[width=1.\linewidth]{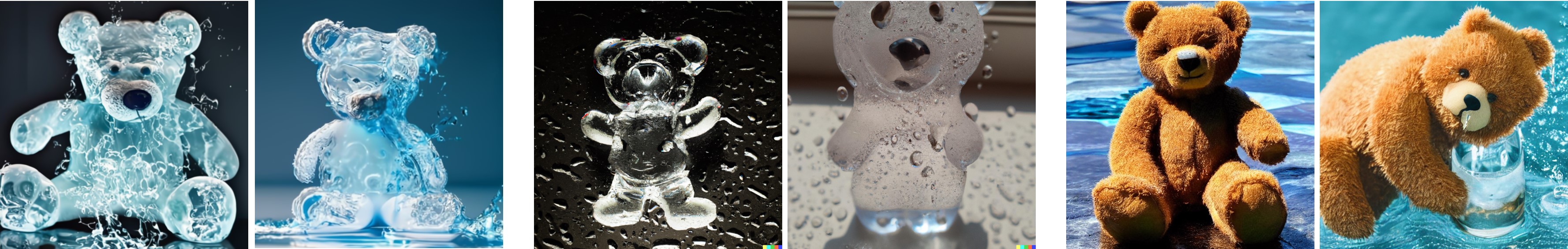}
        \caption{ A photo of a teddy bear made of water.}
    \end{subfigure}       
    \begin{subfigure}{0.95\linewidth}
        \includegraphics[width=1.\linewidth]{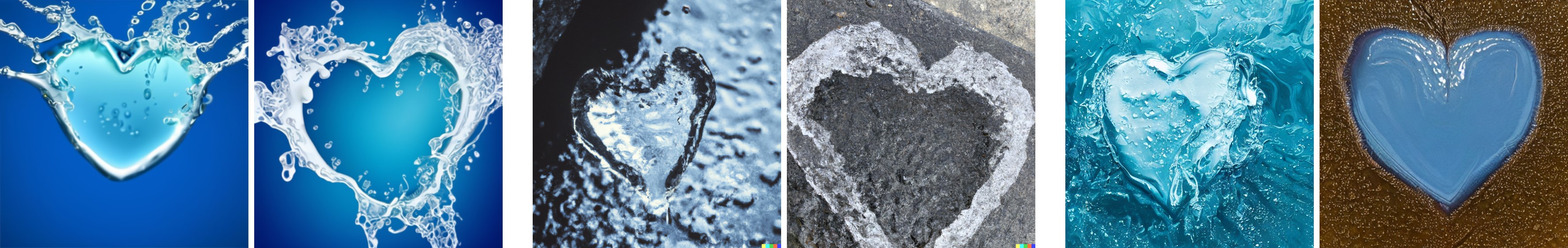} 
        \caption{ A heart made of water.}
    \end{subfigure}          
     \begin{subfigure}{0.95\linewidth}
        \includegraphics[width=1.\linewidth]{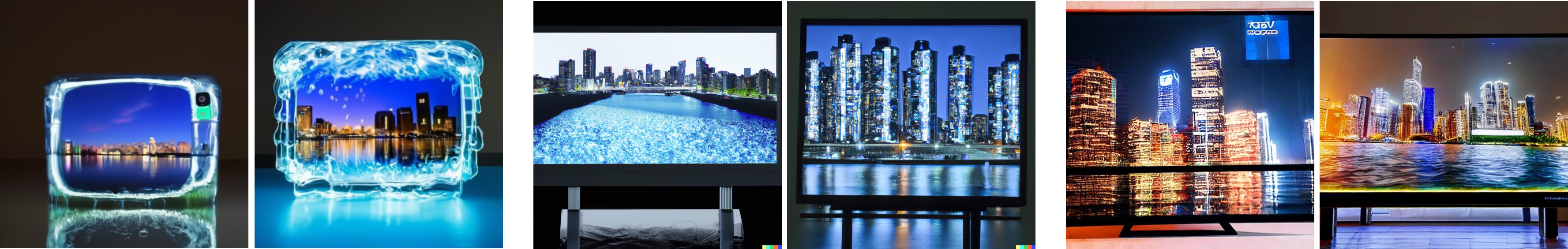}
        \caption{A television made of water that displays an image of a cityscape at night.}
    \end{subfigure}          
    \begin{subfigure}{0.95\linewidth}
        \includegraphics[width=1.\linewidth]{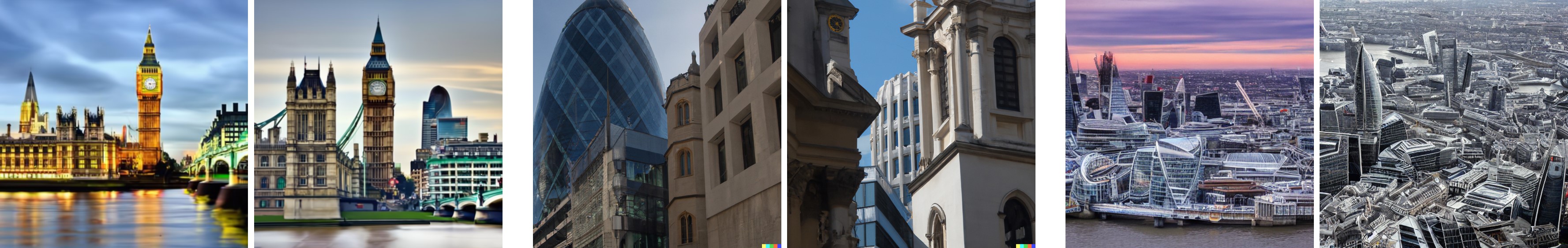}
        \caption{The city of London.}
    \end{subfigure}  
    \begin{subfigure}{0.95\linewidth}
        \includegraphics[width=1.\linewidth]{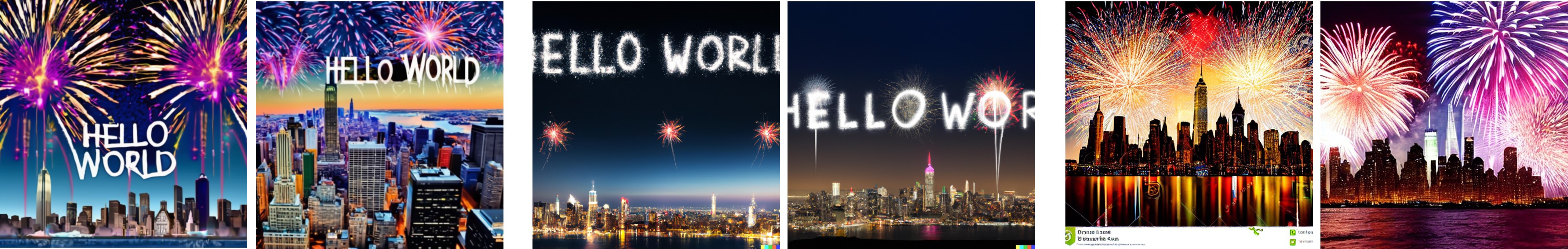}
    \caption{New York Skyline with 'Hello World' written with fireworks on the sky.}
    \end{subfigure}   
    \vspace{-0.15in}
    \caption{Comparison with DALL-E 2 and Stable Diffusion.}
    \label{fig:comparison_2}
\end{figure*}

\begin{figure*}[ht!]
    \centering
    ~~~~~~~~
    \begin{subfigure}{0.8\linewidth}
        \includegraphics[width=1.\linewidth]{figures/Comparison/title.png}
    \end{subfigure}             
    \begin{subfigure}{0.95\linewidth}
        \includegraphics[width=1.\linewidth]{figures/Comparison/c_a_portrait_of_a_statue_of_the_Egyptian_god_Anubis_wearing_av.jpg}
        \caption{A portrait of a statue of the Egyptian god Anubis wearing aviator goggles, white t-shirt and leather jacket. A full moon over the city of Los Angeles is in the background at night.}
    \end{subfigure}        
    \begin{subfigure}{0.95\linewidth}
        \includegraphics[width=1.\linewidth]{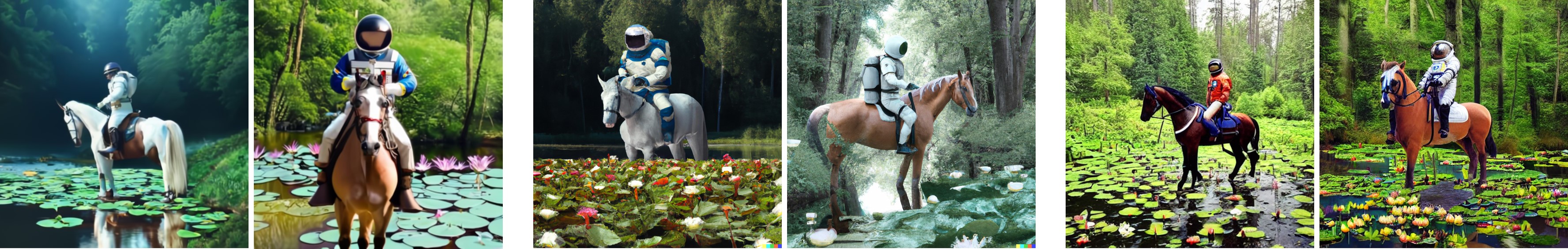}
        \caption{A photo of an astronaut riding a horse in the forest. There is a river in front of them with water lilies.}
    \end{subfigure}       
 
    \begin{subfigure}{0.95\linewidth}
        \includegraphics[width=1.\linewidth]{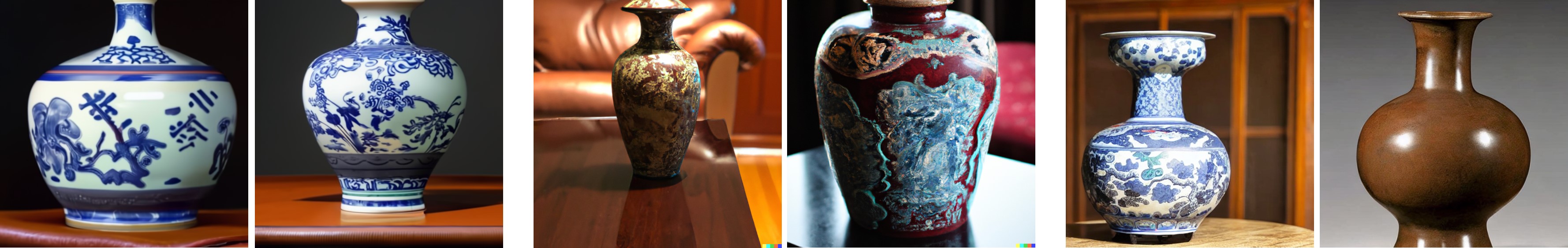}
        \caption{A photo of a Ming Dynasty vase on a leather topped table.}
    \end{subfigure}      
    \begin{subfigure}{0.95\linewidth}
        \includegraphics[width=1.\linewidth]{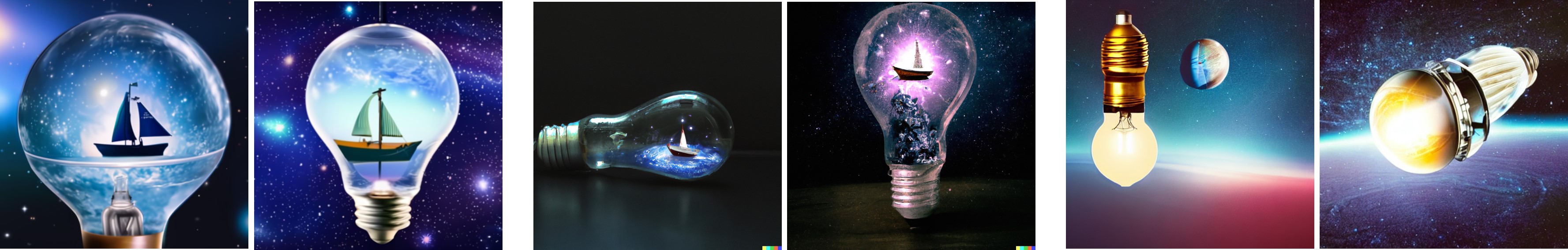}
        \caption{A photo of a light bulb in outer space traveling the galaxy with a sailing boat inside the light bulb.}
    \end{subfigure}     
    \begin{subfigure}{0.95\linewidth}
        \includegraphics[width=1.\linewidth]{figures/Comparison/c_a_statue_of_Abraham_Lincoln_wearing_an_opaque_and_shiny_astronauts_helmet_The_statue_sits_on_the_moon_with_the_planet_Earth_in_the_sky.jpg}
        \caption{ A statue of Abraham Lincoln wearing an opaque and shiny astronaut's helmet. The statue sits on the moon, with the planet Earth in the sky.}
    \end{subfigure}       
    \begin{subfigure}{0.95\linewidth}
        \includegraphics[width=1.\linewidth]{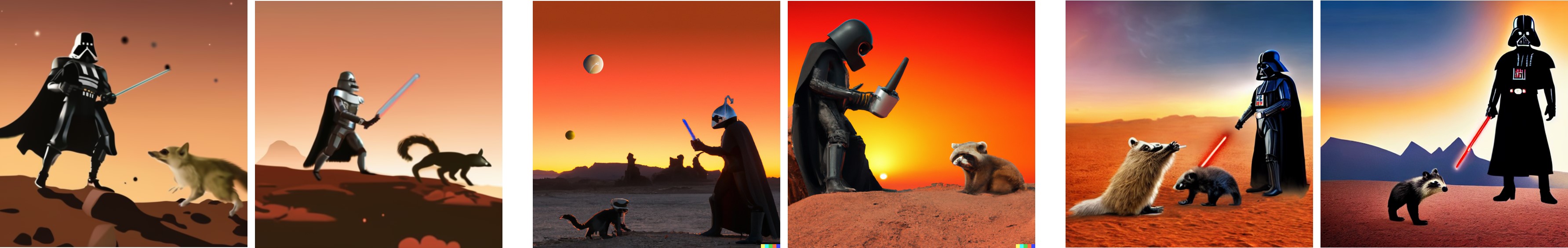}
        \caption{Darth Vader playing with raccoon in Mars during sunset.}
    \end{subfigure}   
    \begin{subfigure}{0.95\linewidth}
        \includegraphics[width=1.\linewidth]{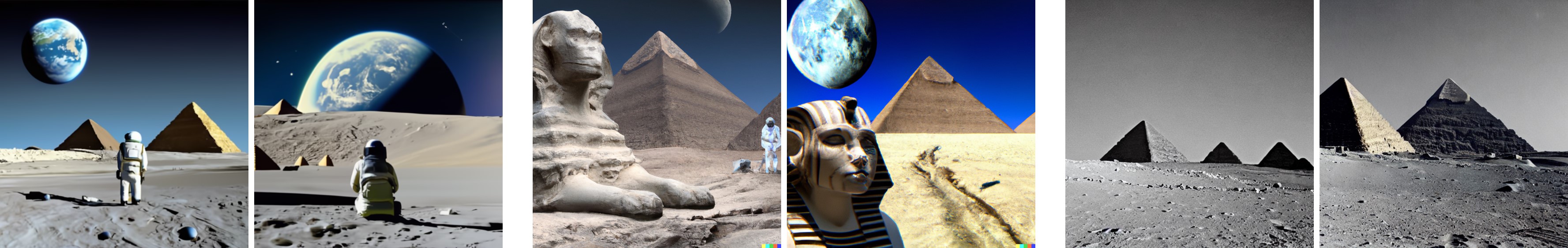}
        \caption{Ground view of the Great Pyramids and Sphinx on the moon's surface. The back of an astronaut is in the foreground. The planet Earth looms in the sky.}
    \end{subfigure}         
    \vspace{-0.25in}
    \caption{Comparison with DALL-E 2 and Stable Diffusion.}
    \label{fig:comparison_3}
\end{figure*}

\begin{figure*}[ht!]
    \centering
    ~~~~~~~~
    \begin{subfigure}{0.8\linewidth}
        \includegraphics[width=1.\linewidth]{figures/Comparison/title.png}
    \end{subfigure}             
    \begin{subfigure}{0.95\linewidth}
        \includegraphics[width=1.\linewidth]{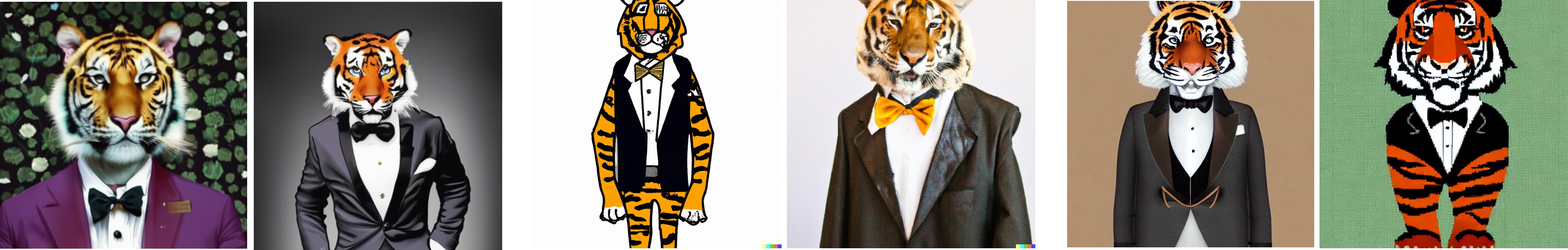}
        \caption{A  tiger wearing a tuxedo.}
    \end{subfigure}        
    \begin{subfigure}{0.95\linewidth}
        \includegraphics[width=1.\linewidth]{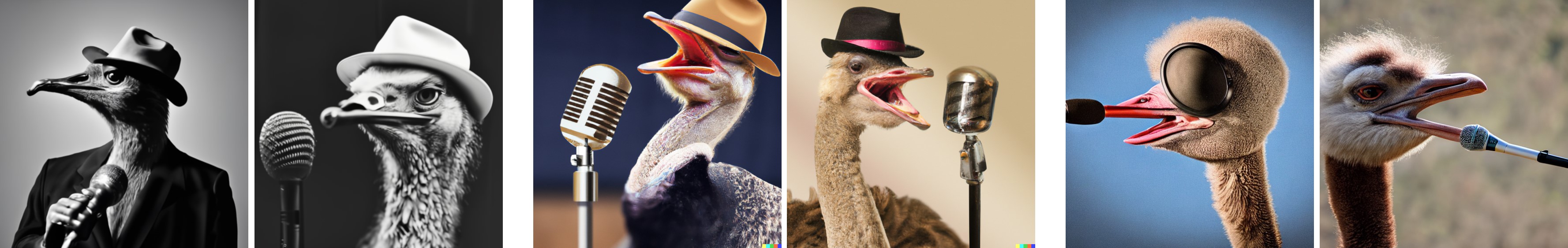} 
        \caption{A photograph of an ostrich wearing a fedora and singing soulfully into a microphone.}
    \end{subfigure}      
    \begin{subfigure}{0.95\linewidth}
        \includegraphics[width=1.\linewidth]{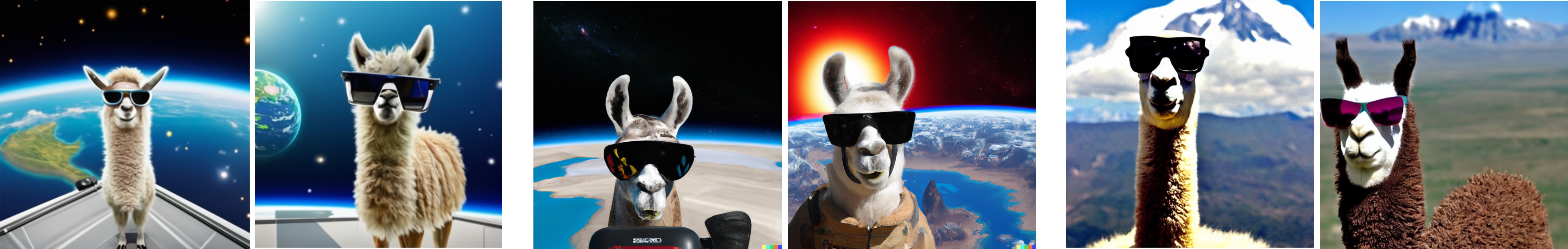}
        \caption{ A photo of llama wearing sunglasses standing on the deck of a spaceship with the Earth in the background.}
    \end{subfigure}   
    \begin{subfigure}{0.95\linewidth}
        \includegraphics[width=1.\linewidth]{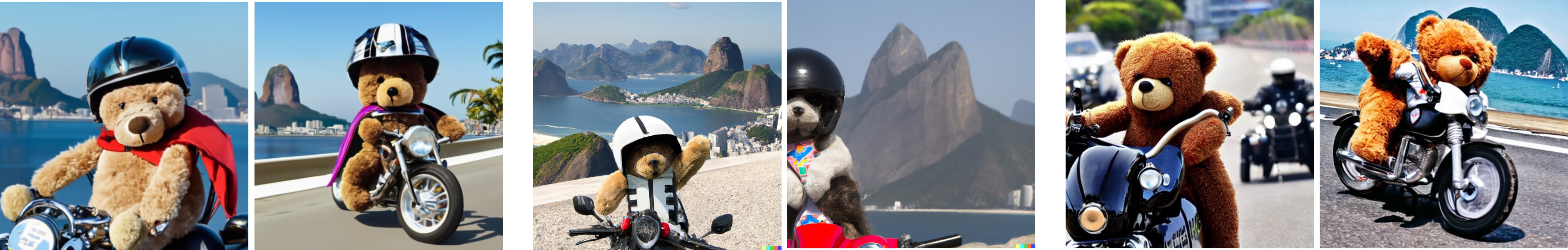}
        \caption{ A teddy bear wearing a motorcycle helmet and cape is riding a motorcycle in Rio de Janeiro with Dois Irmãos in the background.}
    \end{subfigure}   
    \begin{subfigure}{0.95\linewidth}
        \includegraphics[width=1.\linewidth]{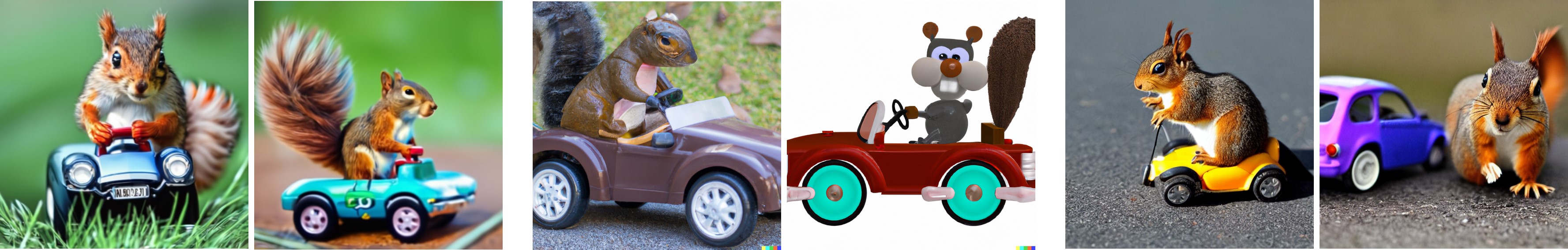}
        \caption{A squirrel driving a toy car.}
    \end{subfigure}       
    \begin{subfigure}{0.95\linewidth}
        \includegraphics[width=1.\linewidth]{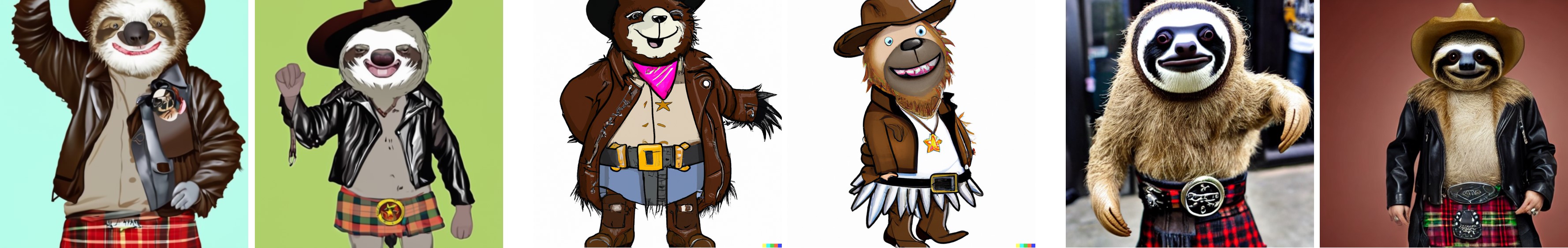}
        \caption{ A smiling sloth wearing a leather jacket, a cowboy hat and a kilt.}
    \end{subfigure}      
    \begin{subfigure}{0.95\linewidth}
        \includegraphics[width=1.\linewidth]{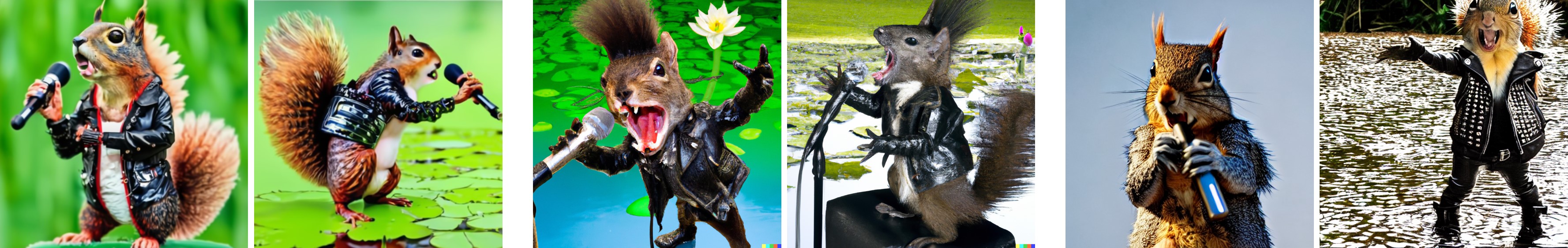}
        \caption{A punk rock squirrel in a studded leather jacket shouting into a microphone while standing on a lily pad.}
    \end{subfigure}       
    \vspace{-0.15in}
    \caption{Comparison with DALL-E 2 and Stable Diffusion.}
    \label{fig:comparison_4}
\end{figure*}

\begin{figure*}[ht!]
    \centering
    ~~~~~~~~
    \begin{subfigure}{0.8\linewidth}
        \includegraphics[width=1.\linewidth]{figures/Comparison/title.png}
    \end{subfigure}            
    \begin{subfigure}{0.95\linewidth}
        \includegraphics[width=1.\linewidth]{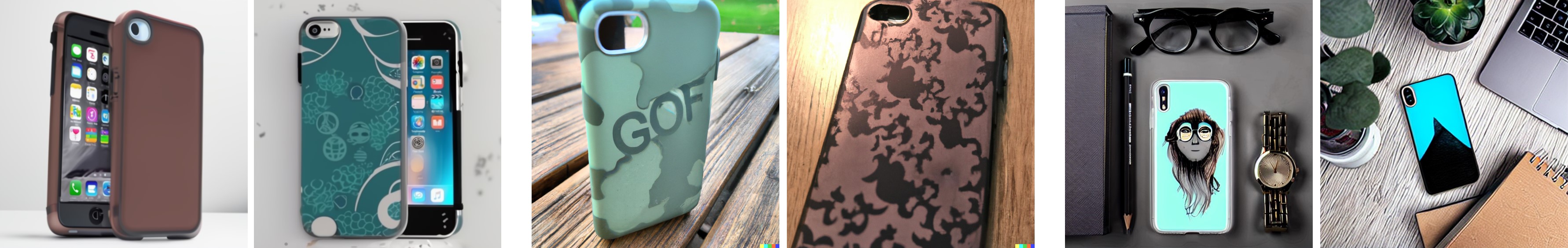}
        \caption{An iPhone case.}
    \end{subfigure}        
    \begin{subfigure}{0.95\linewidth}
        \includegraphics[width=1.\linewidth]{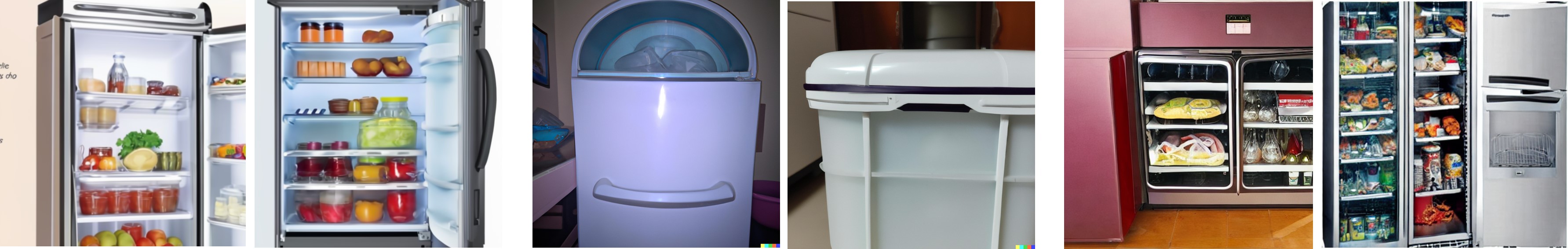}
        \caption{An appliance or compartment which is artificially kept cool and used to store food and drink.}
    \end{subfigure}     
    \begin{subfigure}{0.95\linewidth}
        \includegraphics[width=1.\linewidth]{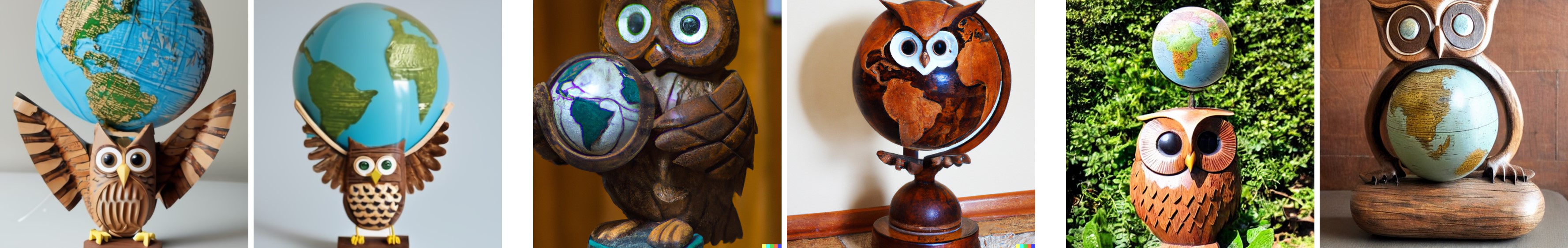}
        \caption{A cute wooden owl statue holding a large globe of the Earth above its head.}
    \end{subfigure}       
    \begin{subfigure}{0.95\linewidth}
        \includegraphics[width=1.\linewidth]{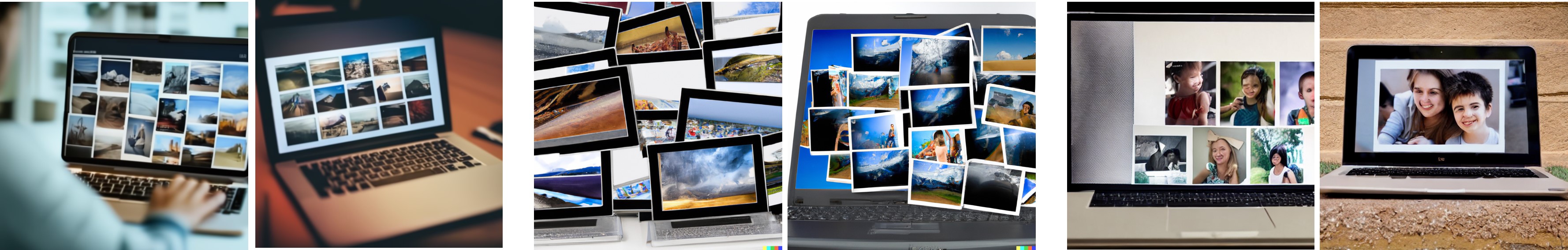}
        \caption{A laptop screen showing a bunch of photographs.}
    \end{subfigure}         
    \begin{subfigure}{0.95\linewidth}
        \includegraphics[width=1.\linewidth]{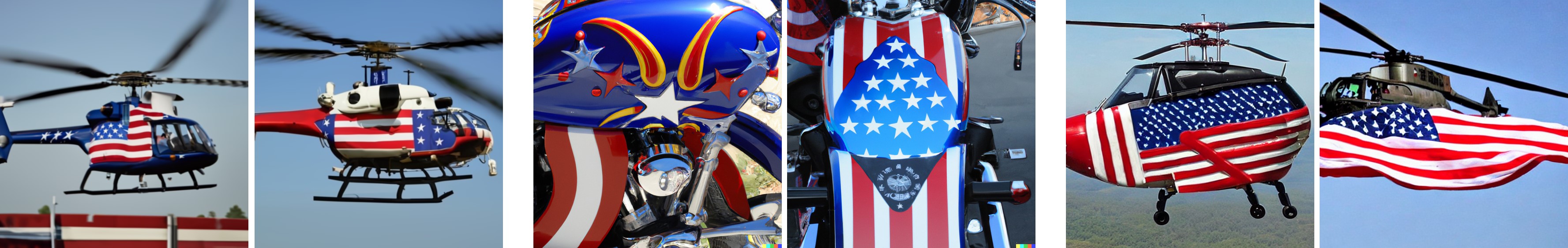}
        \caption{A chopper decorated with the Stars and Stripes..}
    \end{subfigure}       
    \begin{subfigure}{0.95\linewidth}
        \includegraphics[width=1.\linewidth]{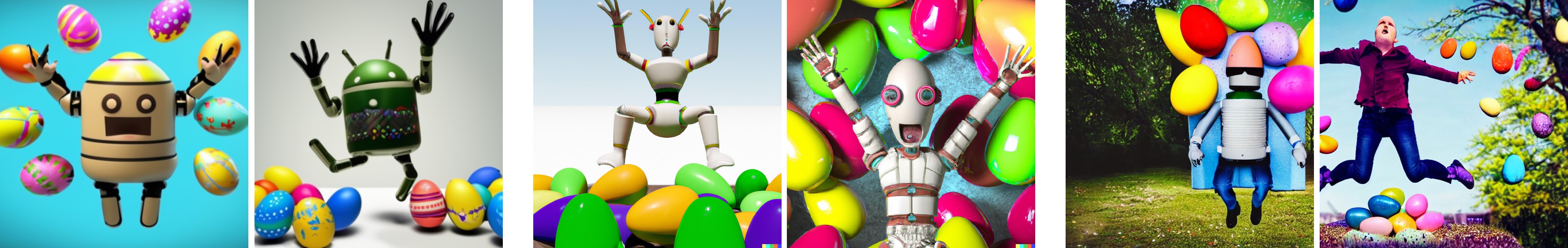}
        \caption{A paranoid android freaking out and jumping into the air because it is surrounded by colorful Easter eggs.}
    \end{subfigure}         
    \begin{subfigure}{0.95\linewidth}
        \includegraphics[width=1.\linewidth]{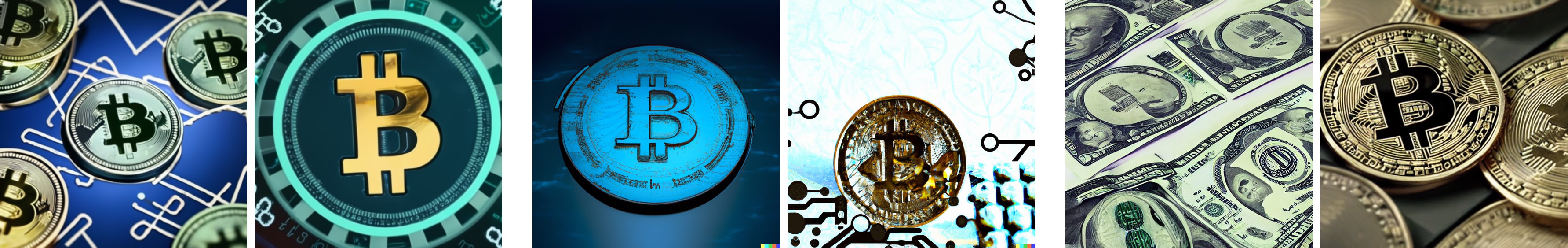}
        \caption{A type of digital currency in which a record of transactions is maintained and new units of currency are generated by the computational solution of mathematical problems, and which operates independently of a central bank.}
    \end{subfigure}        
    \vspace{-0.15in}
    \caption{Comparison with DALL-E 2 and Stable Diffusion.}
    \label{fig:comparison_5}
\end{figure*}

\begin{figure*}[ht!]
    \centering
    ~~~~~~~~
    \begin{subfigure}{0.8\linewidth}
        \includegraphics[width=1.\linewidth]{figures/Comparison/title.png}
    \end{subfigure}             
    \begin{subfigure}{0.95\linewidth}
        \includegraphics[width=1.\linewidth]{figures/Comparison/c_A_map_of_the_United_States_made_out_sushi_It_is_on_a_table_next_to_a_glass_of_red_wine.jpg}
        \caption{A map of the United States made out sushi. It is on a table next to a glass of red wine.}
    \end{subfigure}         
    \begin{subfigure}{0.95\linewidth}
        \includegraphics[width=1.\linewidth]{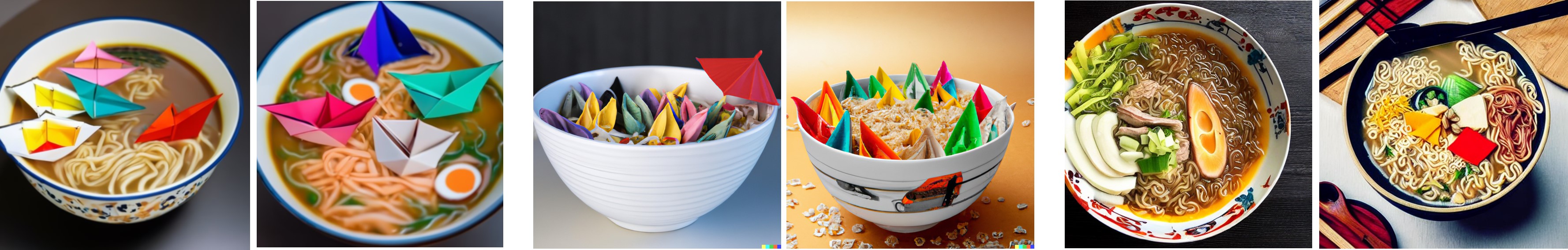}
        \caption{A high resolution photo of a large bowl of ramen. There are several origami boats in the ramen of different colors.}
    \end{subfigure}    
    \begin{subfigure}{0.95\linewidth}
        \includegraphics[width=1.\linewidth]{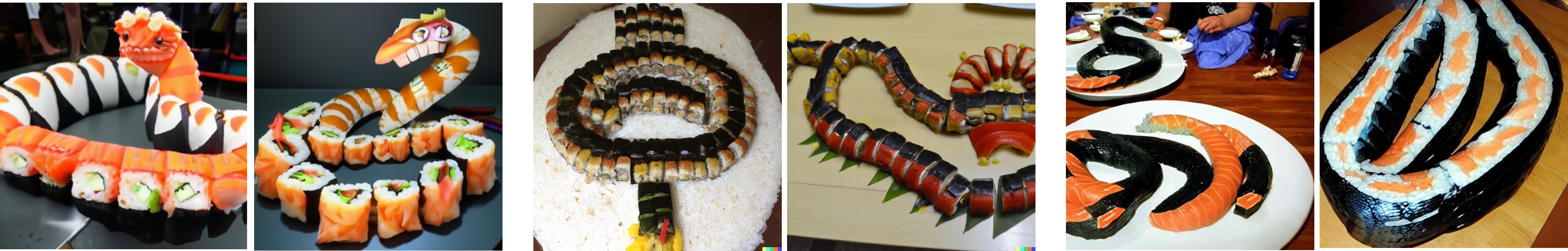}
        \caption{ A giant cobra snake made from sushi.}
    \end{subfigure}    
    \begin{subfigure}{0.95\linewidth}
        \includegraphics[width=1.\linewidth]{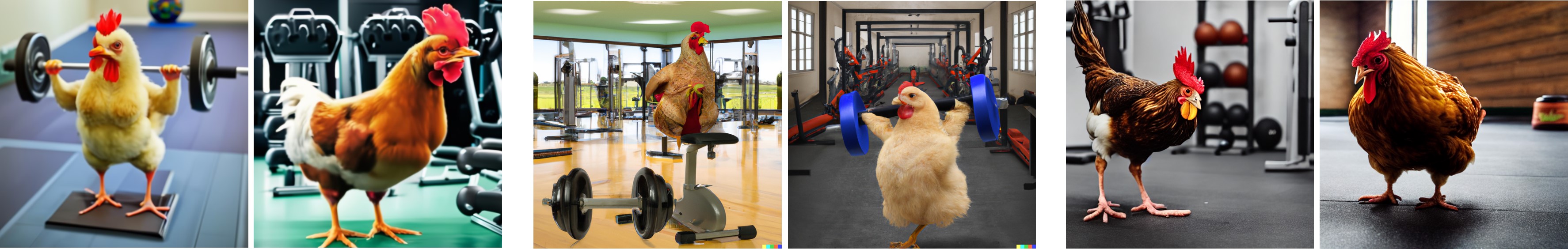}
        \caption{ A high resolution photo of a chicken working out in a gym.}
    \end{subfigure}       
    \begin{subfigure}{0.95\linewidth}
        \includegraphics[width=1.\linewidth]{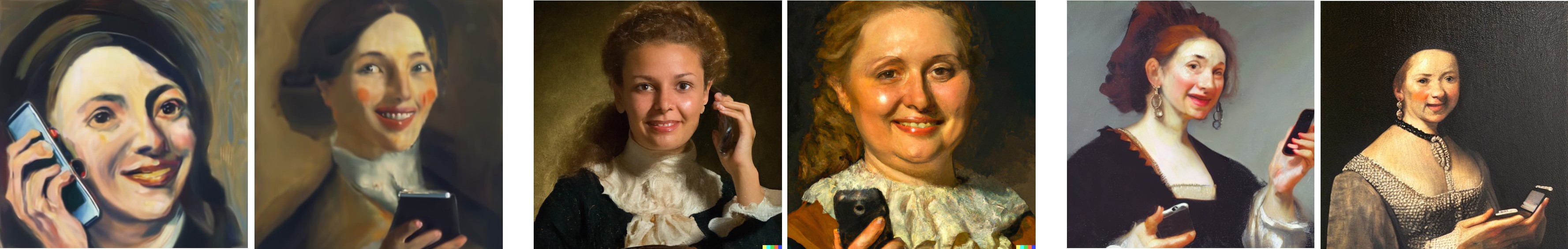}
        \caption{Close-up portrait of a smiling businesswoman holding a cell phone, oil painting in the style of Rembrandt.}
    \end{subfigure}     
    \begin{subfigure}{0.95\linewidth}
        \includegraphics[width=1.\linewidth]{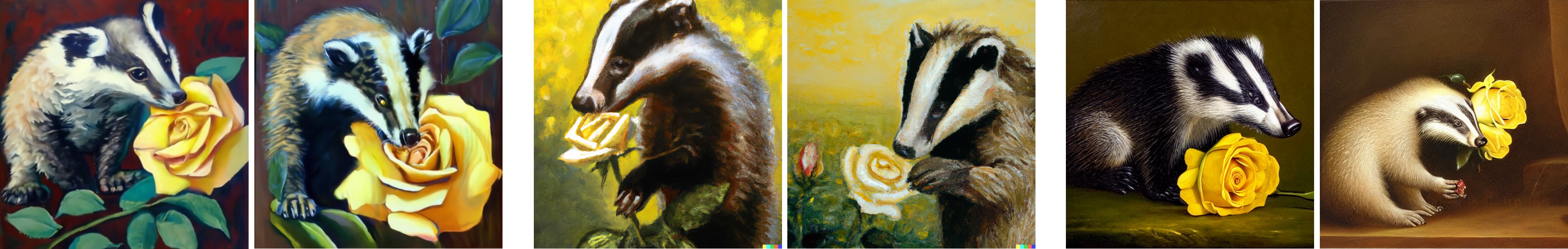}
        \caption{A young badger delicately sniffing a yellow rose, richly textured oil painting.}
    \end{subfigure}      
    \begin{subfigure}{0.95\linewidth}
        \includegraphics[width=1.\linewidth]{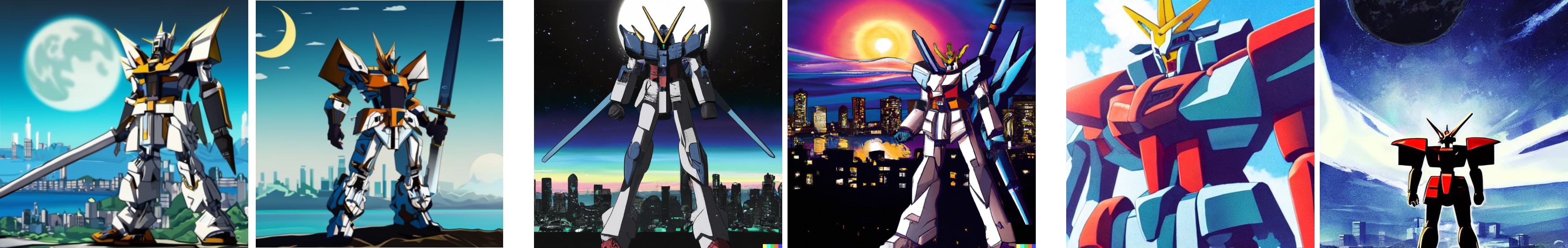}
        \caption{A gundam stands tall with its sword raised. A city with tall skyscrapers is in the distance, with a mountain and ocean in the background. A dark moon is in the sky. realistic high-contrast anime illustration.}
    \end{subfigure}  
    \vspace{-0.15in}
    \caption{Comparison with DALL-E 2 and Stable Diffusion.}
    \label{fig:comparison_6}
\end{figure*}

\begin{figure*}[ht!]
    \centering
    ~~~~~~~~
    \begin{subfigure}{0.8\linewidth}
        \includegraphics[width=1.\linewidth]{figures/Comparison/title.png}
    \end{subfigure}             
    \begin{subfigure}{0.95\linewidth}
        \includegraphics[width=1.\linewidth]{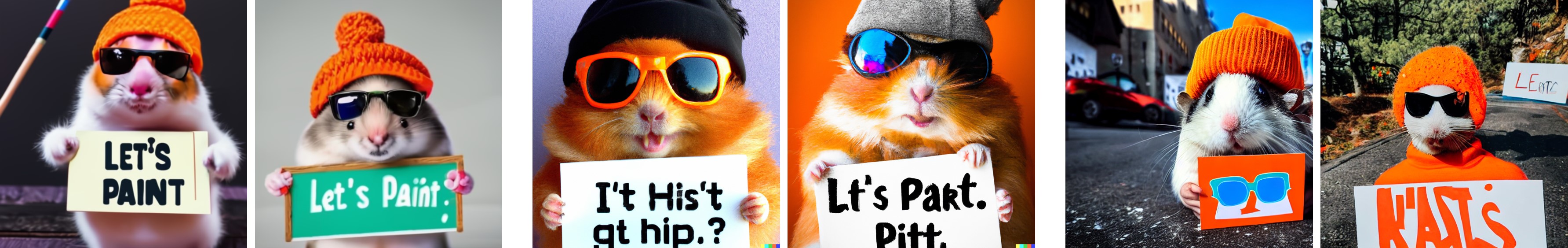}
        \caption{A high contrast portrait photo of a fluffy hamster wearing an orange beanie and sunglasses holding a sign that says Let's PAINT!}
    \end{subfigure}           
    \begin{subfigure}{0.95\linewidth}
        \includegraphics[width=1.\linewidth]{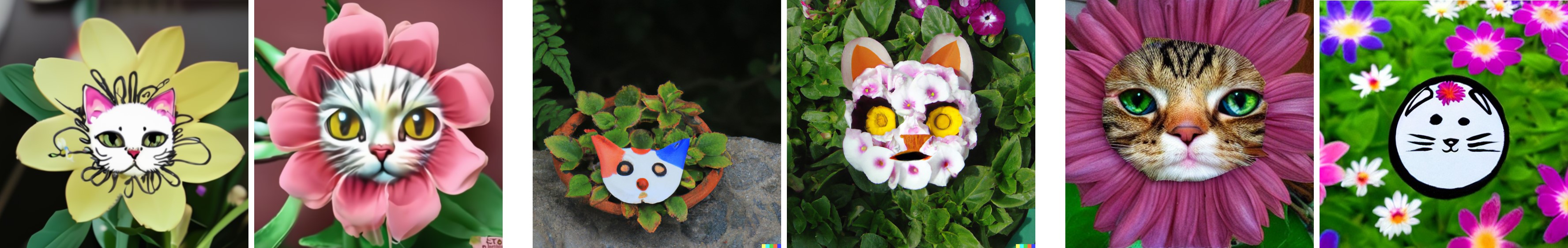}
        \caption{ A  flower with a cat's face in the middle.}
    \end{subfigure}      
    \begin{subfigure}{0.95\linewidth}
        \includegraphics[width=1.\linewidth]{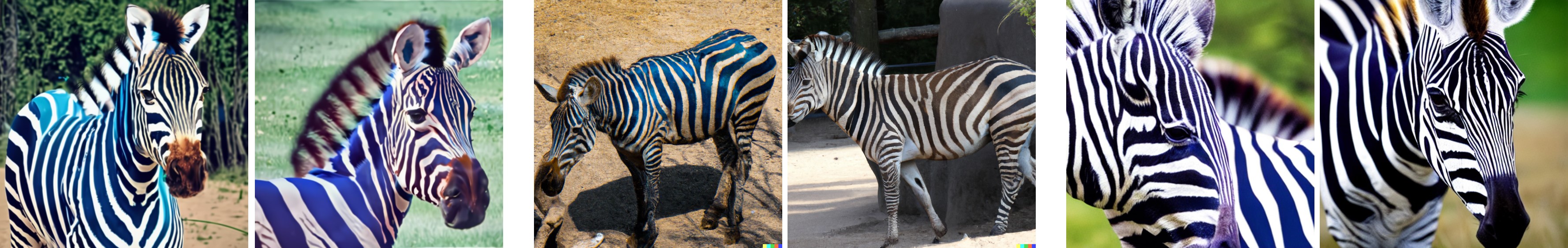} 
        \caption{A zebra with blue stripes.}
    \end{subfigure}      
    \begin{subfigure}{0.95\linewidth}
        \includegraphics[width=1.\linewidth]{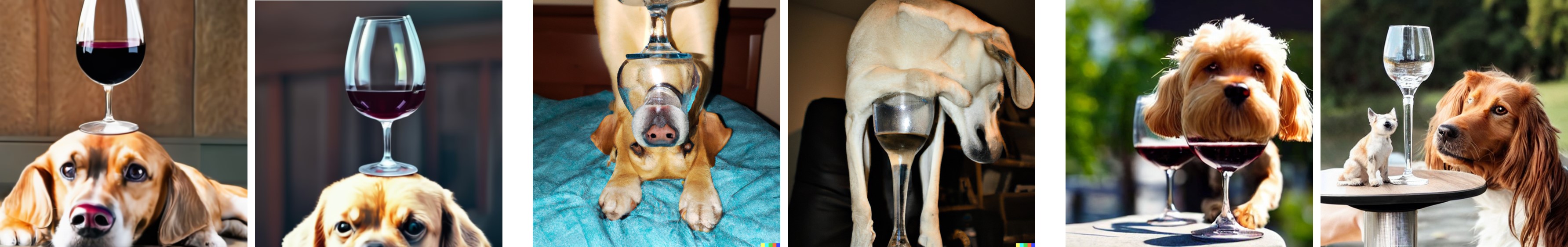}
        \caption{A wine glass on top of a dog.}
    \end{subfigure}        
    \begin{subfigure}{0.95\linewidth}
        \includegraphics[width=1.\linewidth]{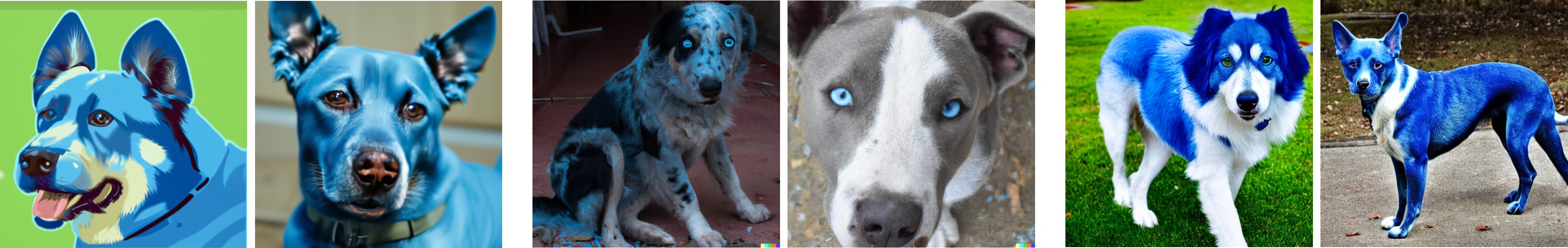}
        \caption{A blue colored dog.}
    \end{subfigure}    
    \begin{subfigure}{0.95\linewidth}
        \includegraphics[width=1.\linewidth]{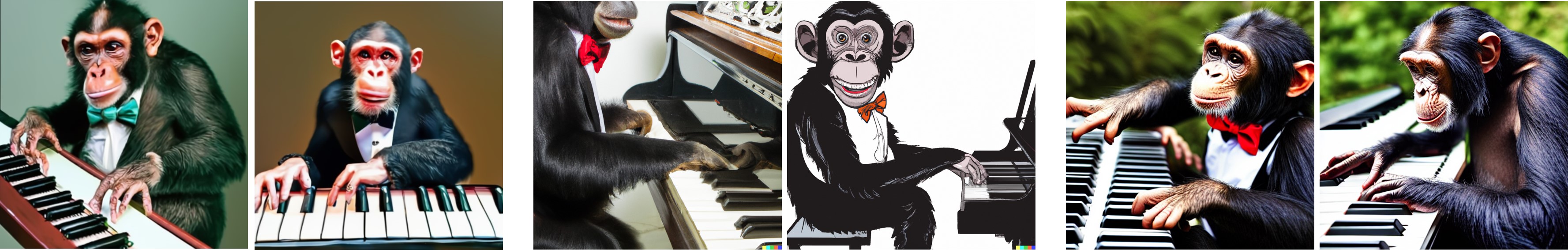}
        \caption{A chimpanzee wearing a bowtie and playing a piano.}
    \end{subfigure}        
    \begin{subfigure}{0.95\linewidth}
        \includegraphics[width=1.\linewidth]{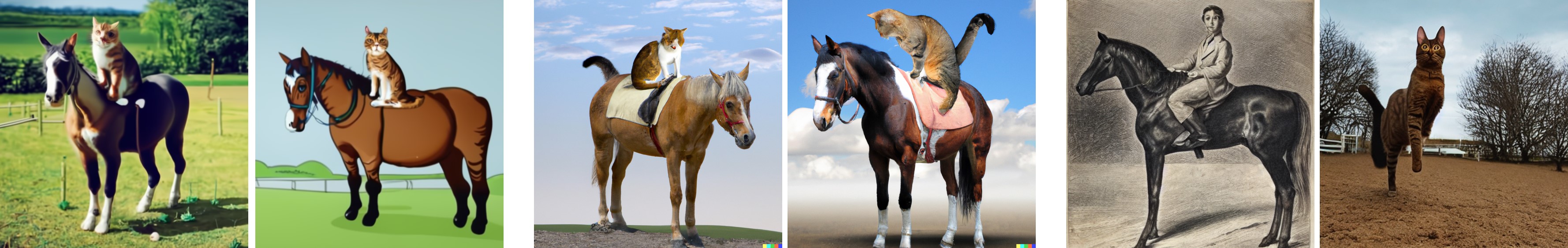}
        \caption{A cat standing on a horse.}
    \end{subfigure}  
    \vspace{-0.15in}
    \caption{Comparison with DALL-E 2 and Stable Diffusion.}
    \label{fig:comparison_7}
\end{figure*}

\end{document}